\documentclass[sn-mathphys-num, iicol]{sn-jnl}% Math and Physical Sciences Numbered Reference Style 
\usepackage{graphicx}%
\usepackage{multirow}%
\usepackage{amsmath,amssymb,amsfonts}%
\usepackage{amsthm}%
\usepackage{mathrsfs}%
\usepackage[title]{appendix}%
\usepackage{xcolor}%
\usepackage{textcomp}%
\usepackage{manyfoot}%
\usepackage{booktabs}%
\usepackage{algorithm}%
\usepackage{algorithmicx}%
\usepackage{algpseudocode}%
\usepackage{listings}%
\usepackage{adjustbox}
\usepackage{float}
\usepackage{placeins}
\usepackage{tabularx}
\usepackage{pifont}
\usepackage{subfigure}
\usepackage{tablefootnote}
\usepackage[font=small]{caption}

\usepackage{makecell}

\usepackage{lipsum}

\usepackage{caption}

\newcommand{\figref}[1]{Fig. \ref{#1}}
\newcommand{\tabref}[1]{Table \ref{#1}}
\newcommand{\equref}[1]{(\ref{#1})}
\newcommand{\secref}[1]{Sec. \ref{#1}}

\newcommand{\cmark}{\ding{51}}%
\newcommand{\xmark}{\ding{55}}%
\begin{document}

\title[Article Title]{Language-guided Recursive Spatiotemporal Graph Modeling for Video Summarization}


\author[1]{\fnm{Jungin} \sur{Park}}\email{newrun@yonsei.ac.kr}

\author*[2]{\fnm{Jiyoung} \sur{Lee}}\email{lee.jiyoung@ewha.ac.kr}

\author*[1, 3]{\fnm{Kwanghoon} \sur{Sohn}}\email{khsohn@yonsei.ac.kr}

% \affil[1]{\orgdiv{School of Electrical and Electronic Engineering}, \orgname{Yonsei University}, \state{Seoul}, \country{Korea}}

% \affil[2]{\orgdiv{NAVER AI Lab}, \orgname{NAVER Corporation}, \orgaddress{\street{Street}, \city{City}, \postcode{10587}, \state{State}, \country{Korea}}}

% \affil[3]{\orgname{Korea Institute of Science and Technology (KIST)}, \orgaddress{\street{Street}, \city{City}, \postcode{610101}, \state{State}, \country{Korea}}}

\affil[1]{\orgname{Yonsei University}, \state{Seoul}, \country{Korea}}

\affil[2]{\orgname{Ewha Womans University}, \state{Seoul}, \country{Korea}}

\affil[3]{\orgname{Korea Institute of Science and Technology (KIST)}, \state{Seoul}, \country{Korea}}


\abstract{
Video summarization aims to select keyframes that are visually diverse and can represent the whole story of a given video. Previous approaches have focused on global interlinkability between frames in a video by temporal modeling. However, fine-grained visual entities, such as objects, are also highly related to the main content of the video. Moreover, language-guided video summarization, which has recently been studied, requires a comprehensive linguistic understanding of complex real-world videos. To consider how all the objects are semantically related to each other, this paper regards video summarization as a language-guided spatiotemporal graph modeling problem. We present recursive spatiotemporal graph networks, called \textit{VideoGraph}, which formulate the objects and frames as nodes of the spatial and temporal graphs, respectively. The nodes in each graph are connected and aggregated with graph edges, representing the semantic relationships between the nodes. To prevent the edges from being configured with visual similarity, we incorporate language queries derived from the video into the graph node representations, enabling them to contain semantic knowledge. In addition, we adopt a recursive strategy to refine initial graphs and correctly classify each frame node as a keyframe. In our experiments, VideoGraph achieves state-of-the-art performance on several benchmarks for generic and query-focused video summarization in both supervised and unsupervised manners.
The code is available at \url{https://github.com/park-jungin/videograph}.}




\maketitle

\section{Introduction}

    Over the years, online video platforms have grown enormously with the development of portable devices and the internet.
    Although this has made a considerable amount of data widely available, video data still needs to be accessed.
    Moreover, the lengths of uploaded videos continue to increase, and it is impractical to watch the videos in full to obtain their valuable information.
    In response to these issues, many recent computer vision researchers have investigated techniques to browse enormous video data efficiently~\cite{tnn-prototype,tnn-reid}.
    In particular, video summarization has become a prominent research topic to automatically create summary videos that describe the original video contents, helping to handle the overwhelming amount of video data~\cite{Zhang16CVPR, Mahasseni17, vs_lstm}.

    \begin{figure*}[!t]
    	\centering
            {\includegraphics[width = 1.0\linewidth]{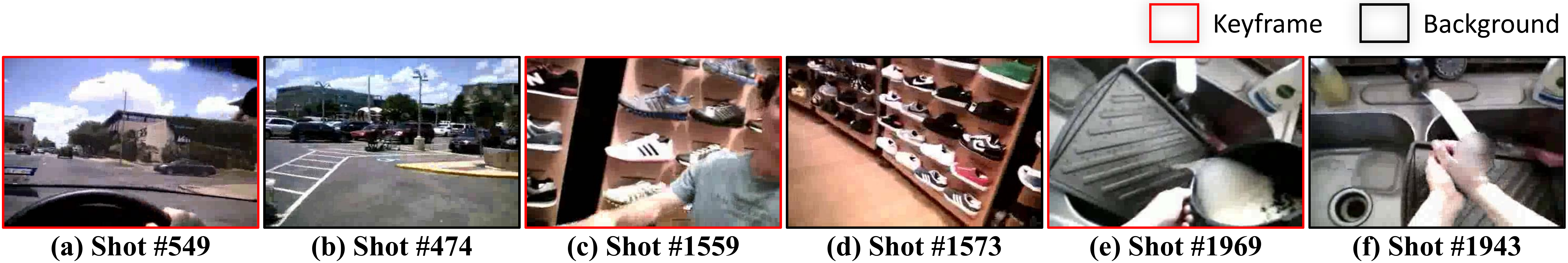}}
     	\caption{Keyframes (red boxes) and background frames (black boxes) from the QFVS benchmark~\cite{qfvs}. To distinguish visually similar frames, we leverage fine-grained visual entities in each frame, enabling a more accurate and robust video summarization.
     	}\label{fig:global}
    \end{figure*}
    
    Variants of recurrent neural networks (RNNs) have been introduced~\cite{vs_lstm,Mahasseni17,He19,Yuan19,hsa-rnn,adsum} to treat video summarization as a sequence labeling problem, achieving substantial accuracy due to their robustness in dealing with sequential data.
    Although RNNs effectively capture long-range dependencies among frames, the signals are progressively propagated through the frames by repeatedly applying recurrent operations.
    Thus, those approaches need to be carefully considered to optimize the models due to this inherent specificity~\cite{he16}.
    They also can cause a multihop dependency modeling problem where delivering messages back and forth between distant positions is difficult~\cite{Wang18nonlocal}.
    Convolutional neural networks (CNNs) based approaches~\cite{Rochan18, Rochan19, park2019video, dsnet}, which have been proposed to tackle these limitations, generally outperform recurrent models.
    However, CNN-based approaches inevitably failed to address semantic relevance between keyframes with varying time distances properly.
    Most recently, transformer~\cite{transformer} has been introduced for video summarization~\cite{vjmht}.
    While they achieved promising results, thanks to the transformer's strong representation capabilities, they require a huge amount of computational cost and a lot of inference time to produce summary videos.
    
    Meanwhile, graphical models have been used specifically to model semantic interactions~\cite{krahenbuhl11,Chen16,kipf2016semi,zeng19,structural-rnn,Park19}, and they have been revisited with graph convolutional networks (GCNs)~\cite{kipf2016semi}, which generalize convolutions from grid-like data to non-grid structures.
    Thus, GCNs have received increasing interest in various computer vision applications, such as object detection~\cite{yuan17, graph-object2, graph-object3}, video classification~\cite{Wang18graph, graph-action2}, video object tracking~\cite{GCT, graph-tracking2}, and action localization~\cite{zeng19, graph-action-local2}.
    These works modeled knowledge graphs based on relationships between different entities, such as images, objects, and temporal proposals.
    Although GCNs have shown promising results, they generally use a fixed graph directly obtained from affinities between feature representations~\cite{Jiang19}, where nodes are strongly connected when they have similar entities.
    Video scene graph generation methods~\cite{vsg1, vsg2, vsg3} have more focused on contextual information for the relationships between objects appearing in the video.
    However, they assumed that the objects, whose relationships are to be inferred, should directly interact with each other.
    Therefore, it is difficult to employ those methods for video summarization directly, which requires identifying comprehensive connections over the whole story to infer the summary video.

    Language-guided video summarization~\cite{clipit,mfst,li23} has been introduced to produce a generic or query-focused summary.
    Language captions generated from the video captioning model~\cite{captionmodel} or user-defined language queries are leveraged into video summarization using transformer~\cite{transformer} and provide customized summaries corresponding to given language queries.
    Despite their flexibility, previous methods~\cite{clipit,mfst,li23} used global visual representations (i.e., frame-level features), leading to inconsistency between visual and language information.
    The language query consists of compositional semantics~\cite{structured-graph,b2a}, while global visual representations contain not only relevant context, but also irrelevant noise~\cite{local-global-context}.
    In practice, as shown in \figref{fig:global}, keyframes ((a), (c), (e)) and background frames ((b), (d), (f)) are often visually similar in global representations but can be distinguished based on fine-grained object details, such as the car handle, the man holding the shoe, and the object held by the hands.
    This inconsistency between the two modality representations makes it difficult to fully utilize the compositional semantics of language queries.
    % we consider that incorporating fine-grained visual representations with language queries can fully utilize the semantic knowledge of language queries.
    In addition, the types of alternative language queries remain underexplored, requiring additional parameters and time for caption generation.

    \begin{figure*}[t]
    \centering
        \renewcommand{\thesubfigure}{}
        	{\includegraphics[width=0.9\linewidth]{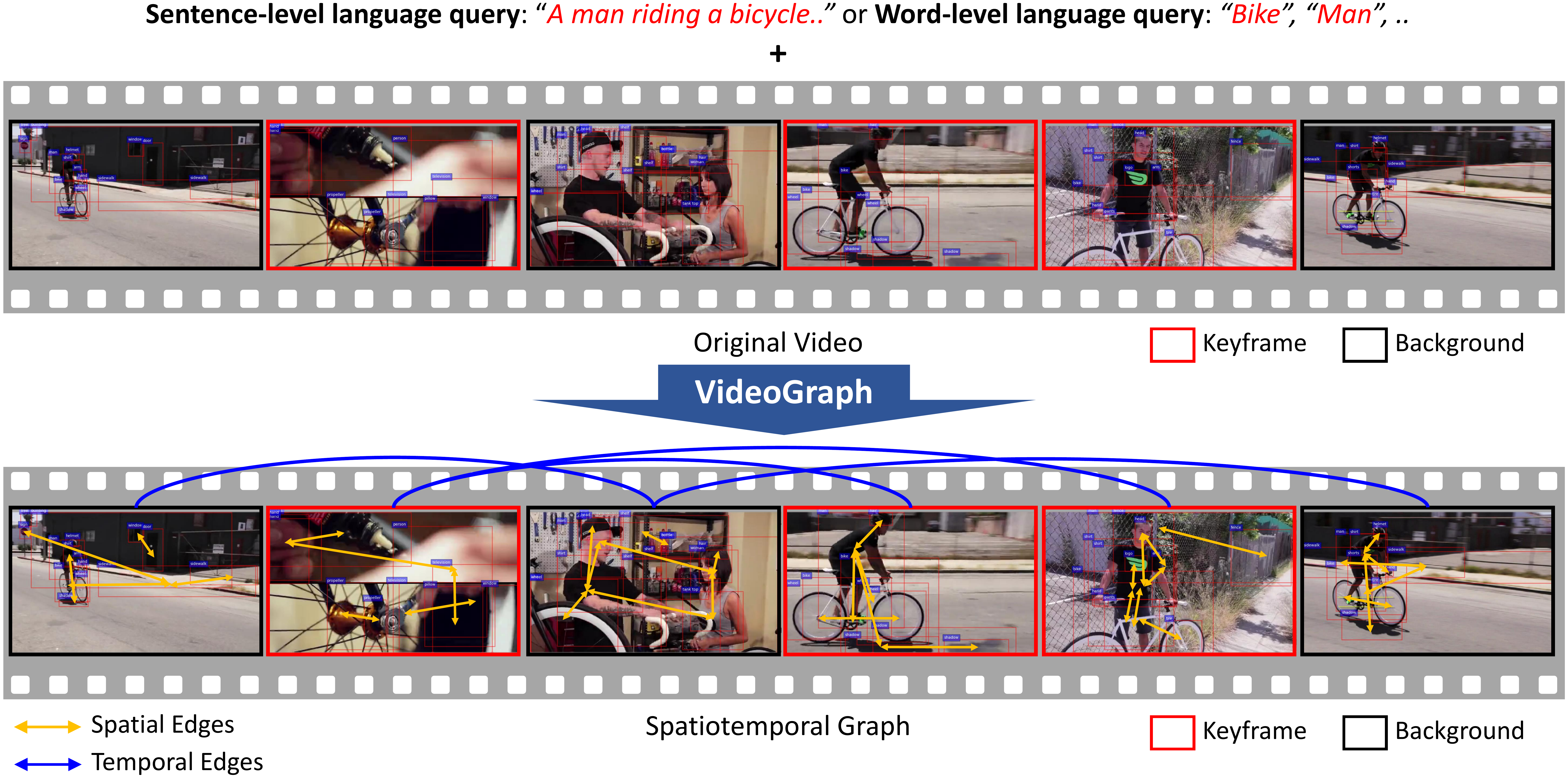}}\hfill\\
        	\caption{
        	Proposed VideoGraph approach, regarding summarization as a language-guided spatiotemporal graph modeling problem.
        	Given video and language queries, richer video summaries are obtained by constructing the language-guided spatiotemporal graphs that represent object-object relationships in each frame and frame-frame relationships in the video.
        }\label{fig:1}
    \end{figure*}
    
    In this paper, we formulate a semantic relationship between frames for video summarization as \textit{a unified relationship linked over time based on the entire video story} and present a novel framework, termed VideoGraph.
    Our VideoGraph incorporates the advantages of deep spatiotemporal graph modeling and a language-guided framework into video summarization by leveraging object-object and frame-frame interactions, as shown in \figref{fig:1}.
    Unlike conventional methods~\cite{vs_lstm,Rochan18,park20,clipit,vjmht} solely using global frame feature representations (e.g. GoogleNet~\cite{szegedy2015going} features), VideoGraph works with object feature representations.
    We formulate a frame as a set of objects within it and construct a spatial graph, where objects are regarded as nodes.
    The object node representations in each frame are connected based on semantic similarity, where the node representations are based on spatially localized features from the object detector~\cite{faster-rcnn} with language features.
    Then the nodes are aggregated along spatial edges through a graph convolution operation.
    The temporal relationships are estimated across different frames by constructing a temporal graph where the averaged object nodes are regarded as nodes.
    The aggregated object representations are propagated frame-to-frame by passing the message through temporal edges, which represent semantic relationships between pairs of frames in the temporal graph.
    
\begin{figure}[t]
            \centering
            	\renewcommand{\thesubfigure}{}
            	\subfigure[(a) Prior works]{\includegraphics[width=0.98\linewidth]{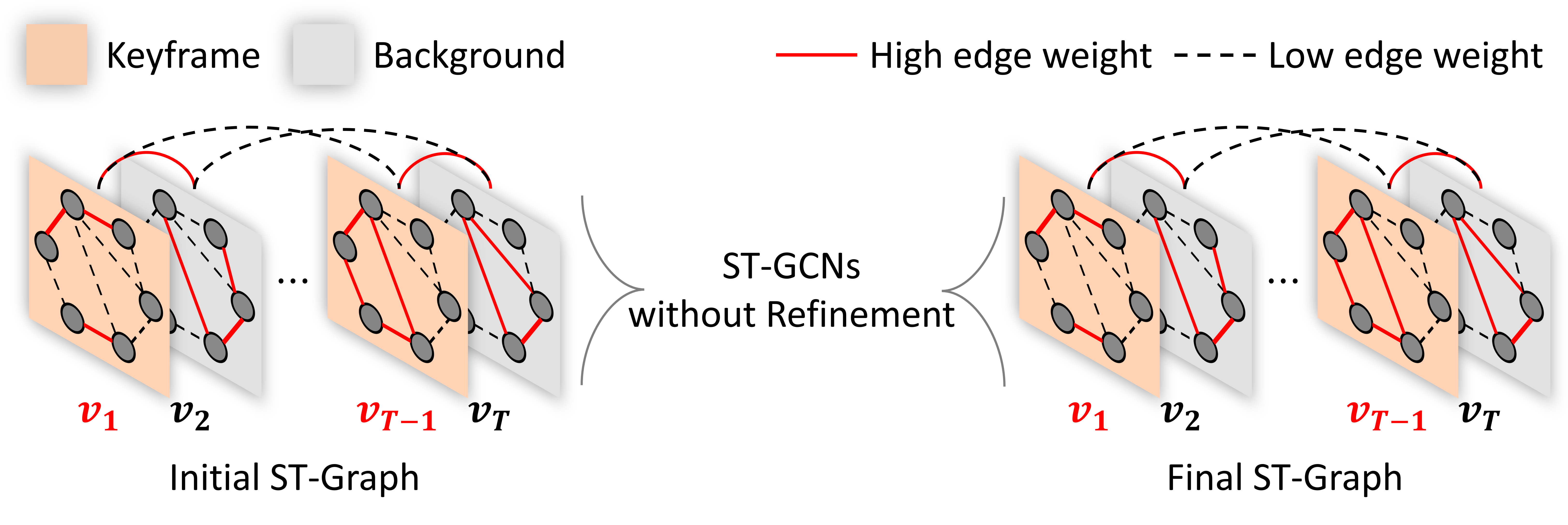}} \\
            	\subfigure[(b) Our approach]{\includegraphics[width=0.98\linewidth]{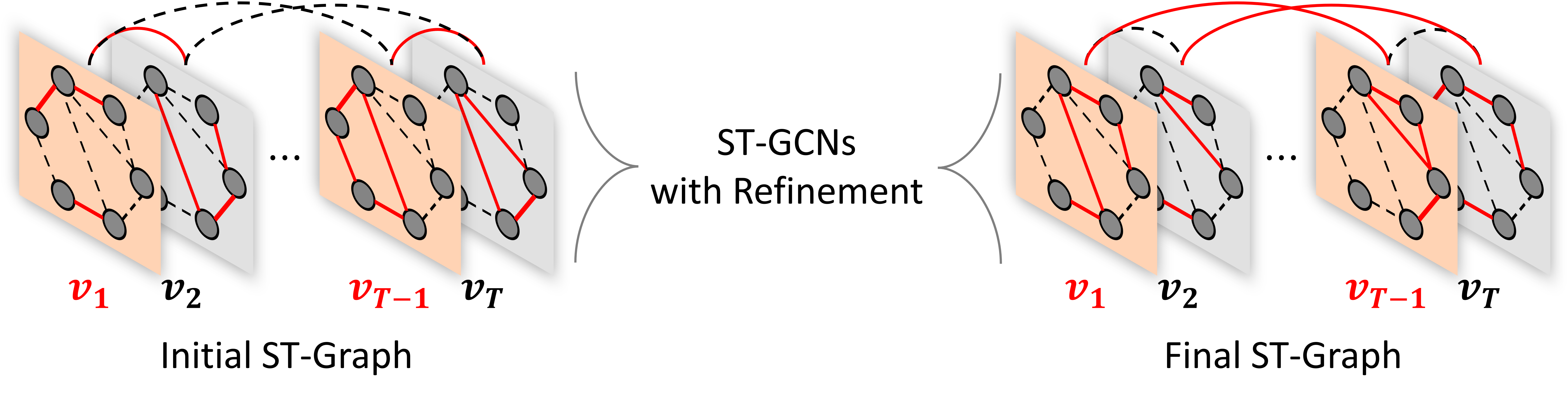}}\hfill
            	\caption{
            	Comparing the proposed VideoGraph with previous ST-GCNs approaches.
            	Circles denote object node representations, red and black squares represent keyframes and backgrounds, respectively:
            	(a) ST-GCNs based methods without graph refinement and (b) VideoGraph refines the initial graph by recursively estimating the graph, hence leveraging semantic relationships.
            }\label{fig:motivation}
        \end{figure}

    In addition, we propose a recursive graph refinement learning scheme that recursively refines the graphs by estimating the residuals of the adjacency matrices to model the semantic similarity (i.e., keyframe or not).
    Unlike the previous works~\cite{GCT,zeng19}, whose constructed relation graphs depend heavily on visual similarity, VideoGraph refines the graphs and enhances feature representations according to the updated graphs~\cite{park20}, as shown in \figref{fig:motivation}.
    At each iteration, the temporal graph is used to estimate the residual spatial adjacency matrix, and the refined spatial graph is used to estimate the residual temporal adjacency matrix to refine the temporal graph.
    With the recursive graph refinement on the language-guided node representations, a frame included in the summary is connected to other keyframe nodes with high affinity weights in the final temporal graph, such that semantic connections between frames can be modeled through the spatial and temporal graphs.
    We extensively examine the proposed method through ablation studies and comparisons with state-of-the-art methods in both supervised and unsupervised manners on the SumMe~\cite{Gygli14}, TVSum~\cite{Song15}, and QFVS~\cite{qfvs} benchmarks.
    
    A preliminary version of this work appeared in~\cite{park20}.
    The current version
    (1) extends the graph topology to the spatiotemporal domain to reason the semantic relationships between fine-grained visual entities (i.e., objects) in the video;
    (2) presents the language-guided graph construction framework to leverage semantic knowledge of language queries and further enhance semantic relationships between the graph nodes by exploring two different types of language queries;
    (3) extends the recursive learning scheme to refine feature representations from spatial to temporal levels; and 
    (4) contains extensive experiments on the SumMe~\cite{Gygli14}, TVSum~\cite{Song15}, and  QFVS~\cite{qfvs} datasets for the generic and query-focused video summarization.

\section{Related Work}

    \subsection{Video Summarization}
    Video summarization condenses a given video into a short synopsis, including video synopses~\cite{pritch07}; time lapse clips~\cite{joshi15, poleg15}; montages~\cite{kang06, sun14}, or storyboards~\cite{gong14, Gygli15}.
    Early video summarization relied primarily on handcrafted criteria~\cite{kang06,lee12,liu02,ngo03,lu13,potapov14}, such as importance, relevance, representativeness, and diversity to produce a summary video.
    
    Deep networks have recently been applied with considerable success~\cite{vs_lstm,Rochan18,Zhang18,Rochan19,park20,hsa-rnn,adsum};
    Recurrent models are widely used to capture the variable range dependencies between frames~\cite{vs_lstm,Mahasseni17,Zhang18,He19,Yuan19,hsa-rnn,adsum,missing1}.
    Although RNN-based approaches have been applied to video data, the recurrent operations are sequential, limiting the simultaneous processing of all the frames.
    To overcome this limitation, CNN-based approaches have been proposed by formulating video summarization as a binary label prediction problem~\cite{Rochan18} and a temporal boundary regression problem~\cite{dsnet}.
    However, the consideration of modeling the relationships among the frames, which provides significant cues as to how best to summarize the video, is not addressed in this approach.
    Transformer~\cite{transformer} has recently been exploited for video summarization~\cite{vjmht}.
    While they successfully incorporated the transformers into a cross-video attention module to take into consideration the semantic dependencies across videos, the high computational requirement of the transformers arises as a new problem.
    
    Some existing methods have been extended into an unsupervised training scheme~\cite{Yuan19,Mahasseni17,Rochan19,He19}.
    Mahasseni et al.~\cite{Mahasseni17} extended an LSTM-based framework with a discriminator network without human-annotated summary videos.
    Their frame selector uses a variational autoencoder LSTM to decode the output for reconstruction through selected frames.
    The discriminator is another LSTM network that learns to distinguish between the input video and its reconstruction.
    Rochan et al.~\cite{Rochan19} trained a network from unpaired data using adversarial learning.
    He et al.~\cite{He19} produced weighted frame features to predict importance scores with attentive conditional generative adversarial networks, whereas Yuan et al.~\cite{Yuan19} proposed a cycle-consistent learning objective to relieve the difficulty of unsupervised learning.
    Although these approaches resolved the problems caused by insufficient data, the locality of recurrent and convolutional operations, which does not directly compute relevance between any two frames, is still problematic.
    To address this problem, Park et al.~\cite{park20} introduced the graph convolutional networks (GCNs) to the video summarization.
    They formulated the video summarization as a temporal graph modeling problem and proposed a recursive graph refinement framework in which keyframes are connected by semantic affinities.
    {Similarly, there has been an attempt to leverage the spatiotemporal graph for video summarization~\cite{era}.
    Although they demonstrated that fine-grained visual entities (i.e., objects) can be an alternative to complement frame information, less progress has been made in constructing the graph with semantic relationships between entities.}

    Most recently, language-guided video summarization methods~\cite{clipit,intentvizor} have been proposed to produce the video summary based on the given language queries.
    They demonstrated the effectiveness of the language-guided video summarization framework by learning the interactions between the video and language queries.
    However, they require manually provided keywords from a benchmark or an additional model and computations for caption generation, which limits their applicability.
    While we share the same spirit with their works, we explore more in-depth studies according to types of language queries with fine-grained visual representations.

    The recursive learning scheme in video summarization has also appeared in \cite{li23}.
    % , which iteratively refined frame scores.
    They proposed to iteratively refine frame scores by enhancing the frame representations based on the frame scores predicted from the previous stage.
    Due to their iterative direct prediction process of subsequent stages, the incorrect biases from an early stage can be propagated to the following stages~\cite{error-feedback, irls}.
    In contrast, we define refinement on semantic connections between nodes within graphs, where interactions between nodes are continuously updated to reflect semantic relationships.
     
    \begin{figure*}[!t]
    	\centering
        {\includegraphics[width = 1\linewidth]{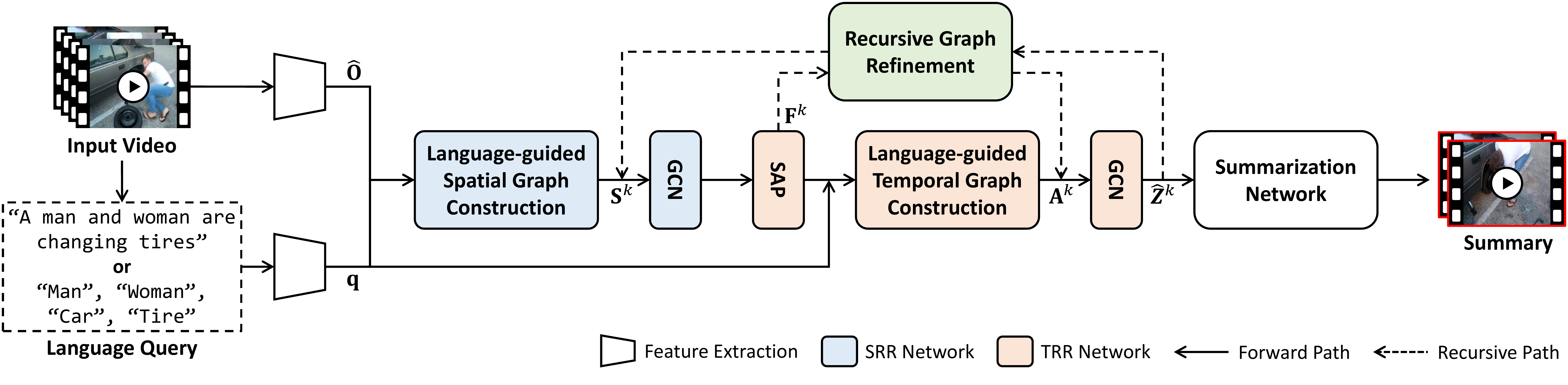}}
     	\caption{
     	Overall framework of VideoGraph comprising feature extraction, spatial relation reasoning (SRR) network, temporal relation reasoning network (TRR), and summarization network in a recurrent structure.
     	}\label{fig:3}
     \end{figure*}
    
    \subsection{Graphical Models}
    Our notion of modeling a graph from a video is partly related to recent research in graphical models.
    One popular direction is using conditional random fields (CRF)~\cite{crf}, especially for semantic segmentation~\cite{krahenbuhl11,Chen16}, where the CRF model is applied to all pairs of pixels in an image to infer mean-field with high confidence.
    
    An attempt at modeling pairwise spatiotemporal relations has been made in non-local neural networks~\cite{Wang18nonlocal}. However, the model is not explicitly defined on graphs. Moreover, the non-local operator is applied to every pixel in the feature space, from lower to higher layers incurring a high computational cost. 
    GCNs~\cite{kipf2016semi} have been used in several computer vision areas, such as video classification~\cite{structural-rnn,Wang18graph,graph-action2}, visual object tracking~\cite{GCT,graph-tracking2}, and temporal action localization~\cite{zeng19,graph-action-local2} to model the temporal relations in a video.
    They have demonstrated the effectiveness of GCNs by exploiting the relations between input data, showing satisfactory performance on each task.
    Similarly, spatiotemporal graph convolutional networks (ST-GCNs)~\cite{st-gcn} have been employed in several computer vision tasks, including video relation detection~\cite{video-relation-detection}, video captioning~\cite{captioning-graph1,captioning-graph2}, spatiotemporal video grounding~\cite{grounding-graph1,grounding-graph2}, emotion recognition~\cite{stgcn-emotion}, person re-identification~\cite{stgcn-reid}, and skeleton-based action recognition~\cite{yan2018spatial,agcn,d-gcn}.
    They have effectively modeled temporal dynamics and spatial dependencies of videos by incorporating spatiotemporal information.
    However, they constructed the spatial graph based on visual and positional similarities between entities~\cite{grounding-graph1,captioning-graph1}. 
    From this perspective, their temporal modeling mainly focused on capturing dynamics in the same entity over time (usually in a few seconds)~\cite{grounding-graph1,grounding-graph2,captioning-graph1,captioning-graph2}.
    In addition, since their graphs are obtained from input feature representations, the connections of graphs may not be optimal~\cite{Jiang19} and are heavily biased to the visual similarities.
    Instead of using fixed initial graphs, we recursively refine the graphs by learning the relationships between nodes to obtain the optimal relation graphs in which edges are connected by a semantic story.
    While some works have attempted to learn the sample-wise spatial graph~\cite{agcn, d-gcn}, their temporal graph is still fixed, accomplishing information propagation along the temporal dimension without considering any interactions between spatial and temporal graphs.
    Contrary to this, the spatial and temporal graphs in VideoGraph are fully trainable and recursively refined from the output of the other graph (e.g. refining the adjacency matrix of the spatial graph from the output of the temporal graph).

    Our work is partially similar to video scene graph generation~\cite{vsg1,vsg2,vsg3} that produces semantic graphs for given videos.
    However, there are several major differences.
    First, they require a complicated object tracking framework to capture the consistently appearing objects in a video.
    Second, they construct graphs for co-occurrence objects that have direct interactions in consecutive frames.
    We simply reuse the pretrained object detector and construct semantic connections for indirectly interacted objects based on the whole story of the video.

\section{Preliminaries}
    This section briefly reviews GCNs~\cite{kipf2016semi}.
    Given a graph $\mathcal{G}$ represented by a tuple $\mathcal{G} = (\mathcal{V}, \mathcal{E})$ where $\mathcal{V}$ is the set of unordered vertices, and $\mathcal{E}$ is the set of edges representing the connectivity between vertices $v \in \mathcal{V}$, GCNs aim to extract richer features at a vertex by aggregating the features from neighboring vertices.
    Suppose vertices $v_i$ and $v_j$ are connected by an edge $e_{ij} \in \mathcal{E}$.
    GCNs represent vertices by associating each vertex $v$ with a feature representation $h_v$.
    The adjacency matrix $\mathbf{A}$ is derived as a $T \times T$ matrix with $A_{ij} = e_{ij}$ representing similarity between vertices $v_i$ and $v_j$ for a fully connected graph, or $A_{ij} = 1$ if $e_{ij} \in \mathcal{E}$, and $A_{ij} = 0$ if $e_{ij} \notin \mathcal{E}$ for a binary graph.

    In standard GCNs, the output of the graph convolution operation is as follows:
    \begin{equation}\label{equ:gcn}
        \mathbf{Z} = \sigma(\mathbf{D}^{-1/2}\hat{\mathbf{A}}\mathbf{D}^{-1/2}\mathbf{X}\mathbf{W}),
    \end{equation}
    where $\sigma(\cdot)$ denotes an activation function such as ReLU, $\mathbf{X}$ and $\mathbf{Z}$ are input and output features, and $\hat{\mathbf{A}} = \mathbf{A} + \mathbf{I}$, where $\mathbf{I}$ is the identity matrix. $\mathbf{D}$ is a diagonal matrix in which a diagonal entry is a sum of the row elements of $\hat{\mathbf{A}}$, and $\mathbf{W}$ is a trainable weight matrix in the graph convolution layer, respectively.
    Given the graph representation, we can perform reasoning on the graph by applying GCNs rather than applying CNNs or RNNs, which have limited capability to represent relationships between features.
    We briefly denote the consecutive graph convolutions as $\mathcal{F}$ in the following sections for the sake of simplicity.

\section{VideoGraph}
    \subsection{Problem Formulation and Overview}
    Inspired by~\cite{ngo03}, VideoGraph learns to model the input video as a graph $\mathcal{G} = \lbrace\mathcal{G}_{S}, \mathcal{G}_{T}\rbrace$ guided by the language query to select a set of keyframes, $\mathcal{K}$, where $\mathcal{G}_{S}$ and $\mathcal{G}_{T}$ are spatial and temporal graphs, respectively.
    The main improvement over \cite{park20} is extending the graph topology to the spatiotemporal domain and leveraging semantic knowledge of language queries.
    The proposed model incorporates the relationships between objects in each frame with edges of the spatial graph and propagates them across frames with edges of the temporal graph.
    Specifically, we regard objects in each frame as nodes and their relationships as edges of the spatial graph, such that the spatial graph can be represented as
    \begin{equation}
        \mathcal{G}_{S} = \lbrace (\mathcal{V}_{S}^{t}, \mathcal{E}_{S}^{t}) \rbrace _{t=1}^{T}, \quad \lbrace v_{S}^{t, n} \rbrace _{n=1}^{N} \subset \mathcal{V}_{S}^{t},
    \end{equation}
    where $\mathcal{V}_{S}^{t}$ and $\mathcal{E}_{S}^{t}$ are the set of nodes and edges in the $t$-th frame, and $v_{S}^{t, n} \in \mathcal{V}_{S}^{n}$ is the $n$-th node in each frame.
    $T$ is the number of video frames and $N$ is the number of objects in each frame.
    The object nodes in each frame are averaged and added by the temporal positional encodings to derive a frame node $v_{t} \in \mathcal{V}_{T}$, and the relationships between frames are used as edges $\mathcal{E}_{T}$ for the temporal graph, such that
    \begin{equation}
        \mathcal{G}_{T} = (\mathcal{V}_{T}, \mathcal{E}_{T}), \quad v_{t} = <\mathcal{V}_{S}^{t}> \oplus \text{PE},
    \end{equation}
    where $v_{t}$ is the $t$-th node of $\mathcal{G}_{T}$, $\oplus$ is the elementwise summation, PE denotes temporal positional encoding, and $<\cdot>$ is the average operation.
    In each graph construction, the node representations are enhanced with a language query representation using a multi-head cross-attention to transfer semantic knowledge of the language query to the node representations.

    In conventional GCNs, the graph is initially constructed from the input data through explicit graphical modeling and fixed in its initial form.
    In contrast to the prior works that capture the visual similarity by using fixed graphs~\cite{GCT, zeng19}, we seek to refine the initial graph iteratively such that the nodes are connected by the story of the video.
    To accomplish this, we formulate the recursive spatiotemporal graph relation networks that gradually complete the semantic graph by iteratively estimating the residuals of the adjacency matrices of the graphs.
    
    As shown in \figref{fig:3}, our networks are split into four parts: 
    1) \textit{feature extraction networks} to extract initial object node representations and a language query representations from pretrained models;
    2) a \textit{spatial relation reasoning network} to build the language-guided spatial graph $\mathcal{G}_{S}$ and estimate the residual temporal adjacency matrix $d\mathbf{A}$, where $\mathbf{A}$ is a temporal adjacency matrix; 
    3) a \textit{temporal relation reasoning network} to construct the language-guided temporal graph $\mathcal{G}_{T}$ from the spatial graph and estimate the residual spatial adjacency matrix $d\mathbf{S}_{t}$, where $\mathbf{S}_{t}$ is a spatial adjacency matrix for the $t$-th frame; and 
    4) a \textit{summarization network} to classify the final frame representations into keyframes for the summary video.

    \subsection{Feature Extraction Network}
    \subsubsection{Visual feature representation}
    In contrast to \cite{park20}, which constructs the temporal relation graph directly from the frame feature representations, we first construct the language-guided spatial graph $\mathcal{G}_S$ in the spatial relation reasoning (SRR) network, where objects in each frame are regarded as nodes and connected by their semantic relationships guided by the language query.
    Although every object node can be connected over frames, we separately construct the spatial graph for each frame to account for computational efficiency.
    Specifically, we extract a set of object feature representations $\mathbf{O}_{t}=\lbrace \mathbf{o}_{t}^{n} \rbrace _{n=1}^{N}$ for the $t$-th frame using an off-the-shelf object detector, where $N$ is the number of objects.
    The extracted feature is fed into a fully-connected (FC) layer to project $\mathbf{o}_{t}^{n}$ to a $d$-dimensional embedding space as follows:
    \begin{equation}
        \hat{\mathbf{o}}_{t}^{n} = \mathbf{W}_{o} \mathbf{o}_{t}^{n} + \mathbf{b}_{o},
    \end{equation}
    where $\hat{\mathbf{O}}_{t}$ is the initial visual representations for the $t$-th frame, $\mathbf{W}_{o}$ and $\mathbf{b}_{o}$ are learnable parameters of the FC layer.

    \subsubsection{Language query representation}
    To explore the effectiveness of language queries, we introduce two different types of language queries. \newline
    \noindent\textbf{Sentence-level language query.}
    We employ an off-the-shelf dense video captioning model~\cite{captionmodel} to generate the sentence corresponding to a given video, following the prior work~\cite{clipit}.
    In this case, we can obtain multiple language captions for each video.
    The generated captions are projected into the feature space through the text encoder (e.g. the pretrained CLIP~\cite{clip} model).
    We concatenate all caption features and feed them into a multi-layer perceptron (MLP) to extract the final language query representation, such that
    \begin{equation}
        \mathbf{q} = \text{MLP}(\text{Concat}(l_1, l_2, ..., l_M)),
    \end{equation}
    where $M$ is the number of generated captions and $l_m$ is the $m$-th caption feature from the text encoder.
    \newline
    \noindent\textbf{Word-level language query.}
    As an alternative to the sentence-level language query, we directly use the object classes detected from the object detector (e.g. car, tire, and person in \figref{fig:3}) as word-level language queries.
    We select $W$ object classes based on the detected frequency over the whole video, assuming that frequently detected objects are closely related to the main contents of the video.
    We embed each word using the text encoder and the MLP layer to low-dimensional embedding space, and project the concatenated word embeddings using the MLP for the final language query representation:
    \begin{equation}
        \mathbf{q} = \text{MLP}(\text{Concat}(c_1, c_2, ..., c_W)),
    \end{equation}
    where $c_w$ is the $w$-th low-dimensional word embedding.

    While the sentence-level language query produces more elaborate descriptions for the video, it requires additional parameters for the dense video captioning model.
    In contrast, the word-level language query can be obtained without any additional captioning model, but it only provides simple keywords.
    We compare the results according to the type of language query in \secref{sec:exp} in terms of the summarization performance, the number of model parameters, and the inference time.

    \begin{figure}[!t]
    	\centering
        	{\includegraphics[width = 1\linewidth]{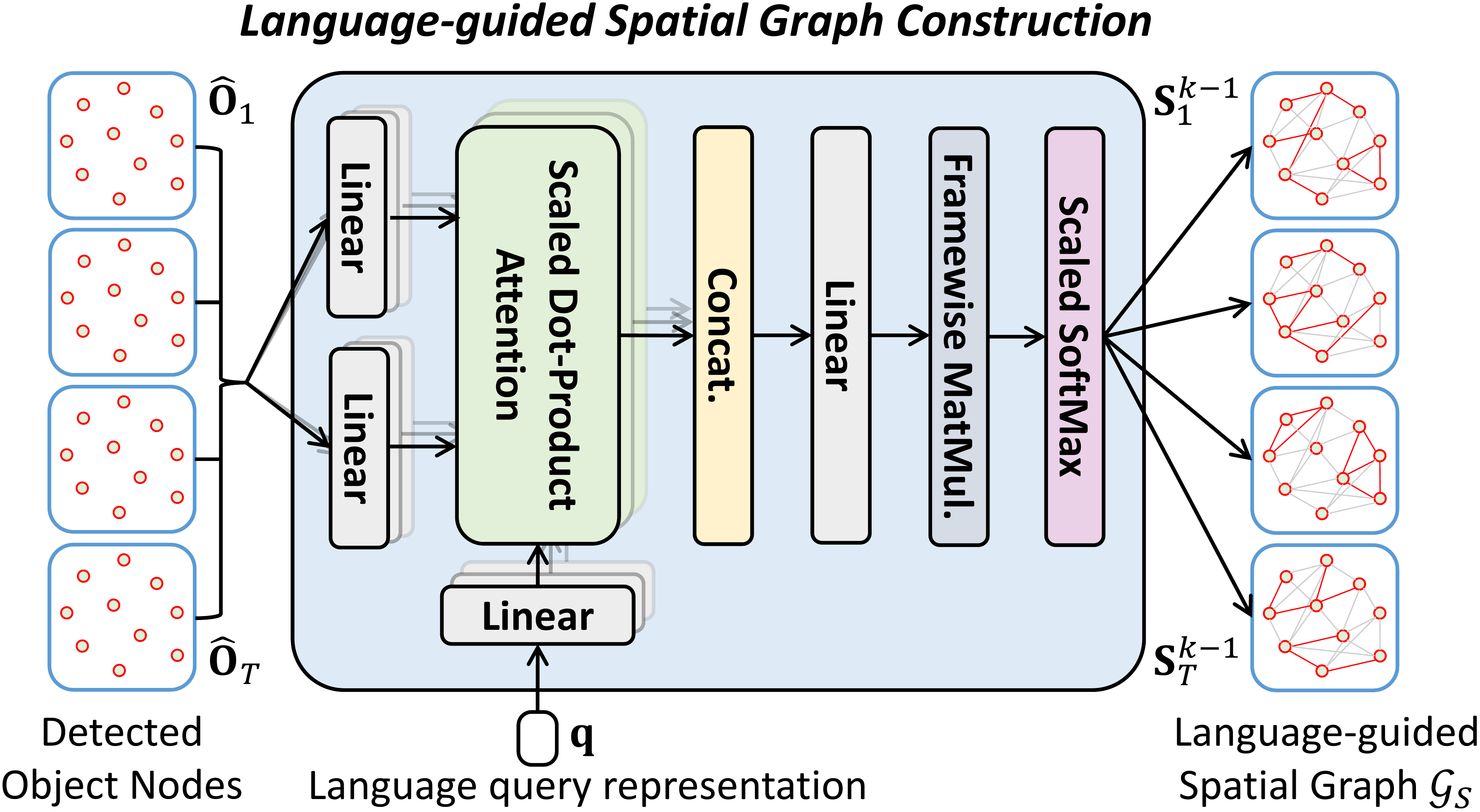}}
     	\caption{
     	Language-guided spatial graph construction module.
     	}\label{fig:4}
    \end{figure}
    
    \subsection{Spatial Relation Reasoning Network}
    We construct the initial language-guided spatial graph $\mathcal{G}_{S}$ for the $t$-th frame as an undirected fully connected graph, as shown in \figref{fig:4}.
    To obtain the set of language-guided object node representations $\tilde{\mathbf{O}}_{t}$, we set the language query  representation $\mathbf{q}$ as a query $Q_S$, and the set of object node representations $\hat{\mathbf{O}}_{t}$ as a key $K_S$ and a value $V_S$ of the modified multi-head cross-attention (MHCA)~\cite{clipit} as follows:
    \begin{equation}
        \begin{gathered}
        Q_S = \mathbf{q},  \quad   K_S, V_S = \hat{\mathbf{O}}_{t},  \\
        \tilde{\mathbf{O}}_{t} = \text{Concat}(\text{head}_1, ..., \text{head}_{H_{S}}) \mathbf{W}_{S},   \\
       \text{head}_h = \text{Attention}(Q_S \mathbf{W}_{h}^{SQ}, K_S \mathbf{W}_{h}^{SK}, V_S \mathbf{W}_{h}^{SV})    \\
        \text{Attention}(Q, K, V) = \text{softmax}(QK^\top/\sqrt{d_{K}})V,            
        \end{gathered}
    \end{equation}
    where $H_S$ denotes the number of heads in the MHCA, $\mathbf{W}_S, \mathbf{W}_h^{SQ}, \mathbf{W}_h^{SK}, \mathbf{W}_h^{SV}$ are the learnable parameter matrices, and $d_{K}$ is the dimension of $K$.
    Different from the prior work~\cite{clipit}, which used the visual features as the query, and the language feature as the key and value, we set the language feature as the query to learn the \textit{language-guided visual representations}.
    The spatial adjacency matrix $\mathbf{S}_{t}$, which represents connections of $\tilde{\mathbf{o}}_t$ and their edge weights, is obtained by computing the affinity between language-guided node representations,
    \begin{equation}\label{equ:S}
        s_{t, ij} = \frac{\exp(\lambda_{o} (\tilde{\mathbf{o}}_{t}^{i})^{\top}\tilde{\mathbf{o}}_{t}^{j})}{\sum_{j=1}^{N} \exp(\lambda_o (\tilde{\mathbf{o}}_{t}^{i})^{\top}\tilde{\mathbf{o}}_{t}^{j})},
    \end{equation}
    where $s_{t, ij}$ indicates the edge weight between the $i$-th and $j$-th nodes in the $t$-th spatial graph, and $\lambda_o$ is a scaling factor.
    
    The language-guided object node representations and the adjacency matrix for the $t$-th frame are fed into consecutive graph convolution layers to aggregate each representation to its neighbor nodes along spatial edges, such that the output of the SRR network is derived by the following equation:
    \begin{equation}\label{equ:Z_S}
        \mathbf{Z}_{t} = \mathcal{F}(\tilde{\mathbf{O}}_{t}, \mathbf{S}_{t}|\mathbf{W}_{z}),
    \end{equation}
    where $\mathbf{Z}_{t}$ is the output of the SRR network for the $t$-th frame and $\mathcal{F}$ is the feed-forward process for consecutive graph convolution layers with the set of learnable parameters $\mathbf{W}_{z}$.
    The aggregated object node representations $\lbrace \mathbf{Z}_{t} \rbrace_{t=1}^{T}$ are used to construct an initial language-guided temporal graph at the first feed-forward pass and estimate the residual temporal adjacency matrix in the recursive graph refinement process.

    \begin{figure}[!t]
    	\centering
        	{\includegraphics[width = 0.93\linewidth]{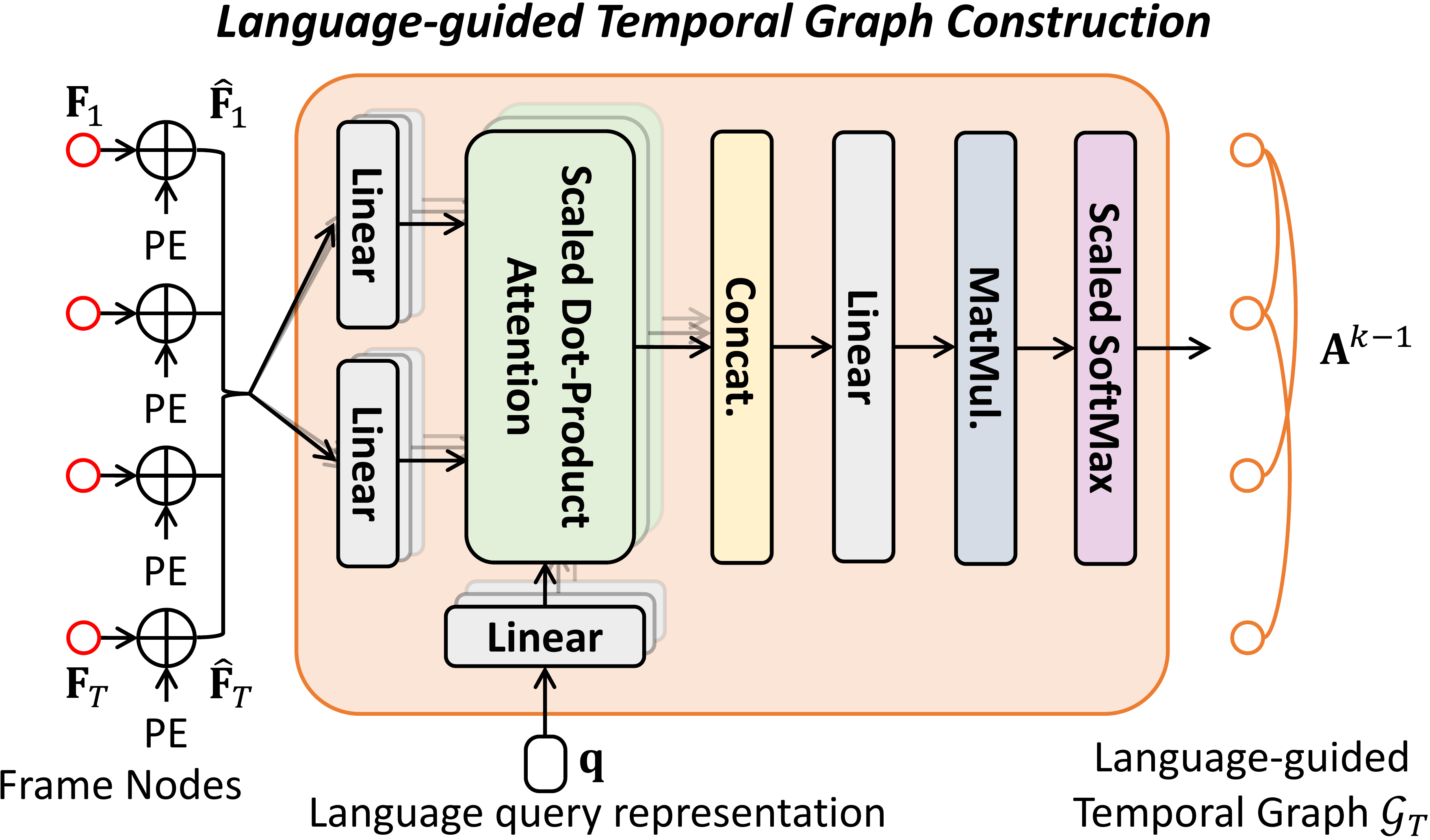}}
     	\caption{
     	Language-guided temporal graph construction module.
     	}\label{fig:5}
    \end{figure}

    \subsection{Temporal Relation Reasoning Network}
    Although relationships between entities in the same frame can be learned in the SRR network, temporal relations that represent semantic relationships between frames are not captured.
    To take temporal dependencies of the video frames into account, we construct the temporal graph $\mathcal{G}_{T}$ in the temporal relation reasoning (TRR) network in which language-guided frame representations are regarded as nodes and fully connected by their semantic affinities, as shown in \figref{fig:5}.
    Concretely, the frame representation is derived from the output of the SRR network by averaging the aggregated object node representations in a corresponding frame, such that the $t$-th frame node representation can be represented as
    \begin{equation}\label{equ:f_F}
        \mathbf{f}_{t} = \frac{1}{N} \sum_{n=1}^{N} {\mathbf{z}_{t}^{n}},
    \end{equation}
    where $N$ is the number of objects in the $t$-th frame.
    Similar as in the SRR network, we set $\mathbf{q}$ as the query, and the frame representations with the positional encoding (PE) as the key and value of the additional MHCA to obtain the language-guided frame node representations $\hat{\mathbf{F}}$:
    \begin{equation}
        \begin{gathered}
        Q_T = \mathbf{q},  \quad   K_T, V_T = \mathbf{F} + \text{PE},  \\
        \hat{\mathbf{F}} = \text{Concat}(\text{head}_1, ..., \text{head}_{H_{T}}) \mathbf{W}_{T}, \\ 
        \text{head}_h = \text{Attention}(Q_T \mathbf{W}_{h}^{FQ}, K_T \mathbf{W}_{h}^{FK}, V_T \mathbf{W}_{h}^{FV}),            
        \end{gathered}
    \end{equation}
    where $H_T$ denotes the number of heads, $\mathbf{W}_{h}^{FQ}, \mathbf{W}_{h}^{FK},$ and $\mathbf{W}_{h}^{FV}$ are the learnable parameters to embed the query, key, and value, respectively.
    The temporal adjacency matrix $\mathbf{A}$ is obtained in the same way as in \equref{equ:S}, and the edge weight $a_{ij}$ between the $i$-th and $j$-th frame nodes is
    \begin{equation}\label{equ:A}
        a_{ij} = \frac{\exp(\lambda_f \hat{\mathbf{f}}_{i}^{\top} \hat{\mathbf{f}}_{j})}{\sum_{j=1}^{T}\exp(\lambda_f \hat{\mathbf{f}}_{i}^{\top} \hat{\mathbf{f}}_{j})},
    \end{equation}
    where $\lambda_{f}$ is a scaling factor for the temporal graph.
    
    Although the frame node representation can be directly propagated to its neighbor nodes through graph convolution, the conventional node propagation cannot be applied to the recursive graph refinement formulation that derives the residual spatial adjacency matrix from frame node representations, since the average operation in \equref{equ:f_F} discards the object coordinates.
    Therefore, we reformulate the graph convolution operation in the TRR network to reason temporal relationships by propagating frame nodes to object nodes over frames along temporal edges rather than through frame-to-frame propagation.
    Formally, the $n$-th object node representation in the $t$-th frame after applying a single graph convolution layer in the TRR network can be expressed as
    \begin{equation}\label{equ:Z_hat}
    \begin{split}
        \hat{\mathbf{z}}_{t}^{n} & = \sigma(\mathbf{W}_{f}(\mathbf{z}_{t}^{n} + \sum_{k\in\mathcal{N}_{t}}\frac{a_{tk}}{N}\sum_{m=1}^{N}\mathbf{z}_{k}^{m})+\mathbf{b}_{f}) \\
                            & = \sigma(\mathbf{W}_{f}(\mathbf{z}_{t}^{n} + \sum_{k\in\mathcal{N}_{t}}a_{tk}\hat{\mathbf{f}}_{k}) + \mathbf{b}_{f}),
    \end{split}
    \end{equation}
    where $\mathbf{W}_{f}$ and $\mathbf{b}_{f}$ are learnable parameters of the graph convolution layer, and $\mathcal{N}_{t}$ is a set of neighboring nodes to the $t$-th frame node.
    This form of aggregation not only maintains the advantages of the graph convolution operation but also retains the object coordinates, allowing the residual spatial adjacency matrix to be estimated from frame nodes.
    The updated object node representations are the d to estimate the residual spatial adjacency matrix $d\mathbf{S}$ in the recursive graph refinement process and fed into the summarization network for graph reasoning after the final iteration.
    
     \begin{figure*}[t]
        	\centering
        	\renewcommand{\thesubfigure}{}
        	\subfigure[(a) Iteration 1]{\includegraphics[width=0.323\textwidth]{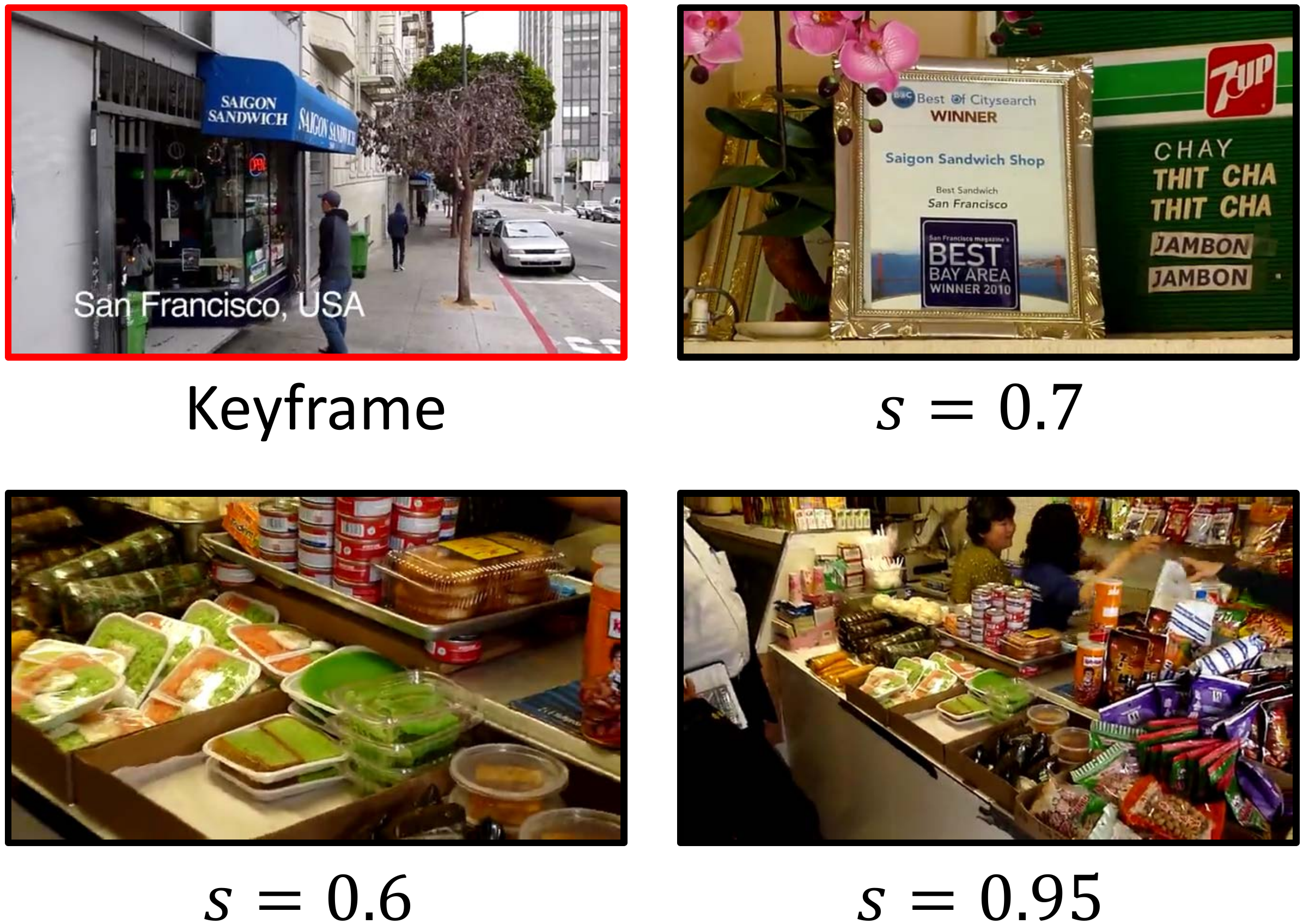}}\hfill 
        	\subfigure[(b) Iteration 3]{\includegraphics[width=0.323\linewidth]{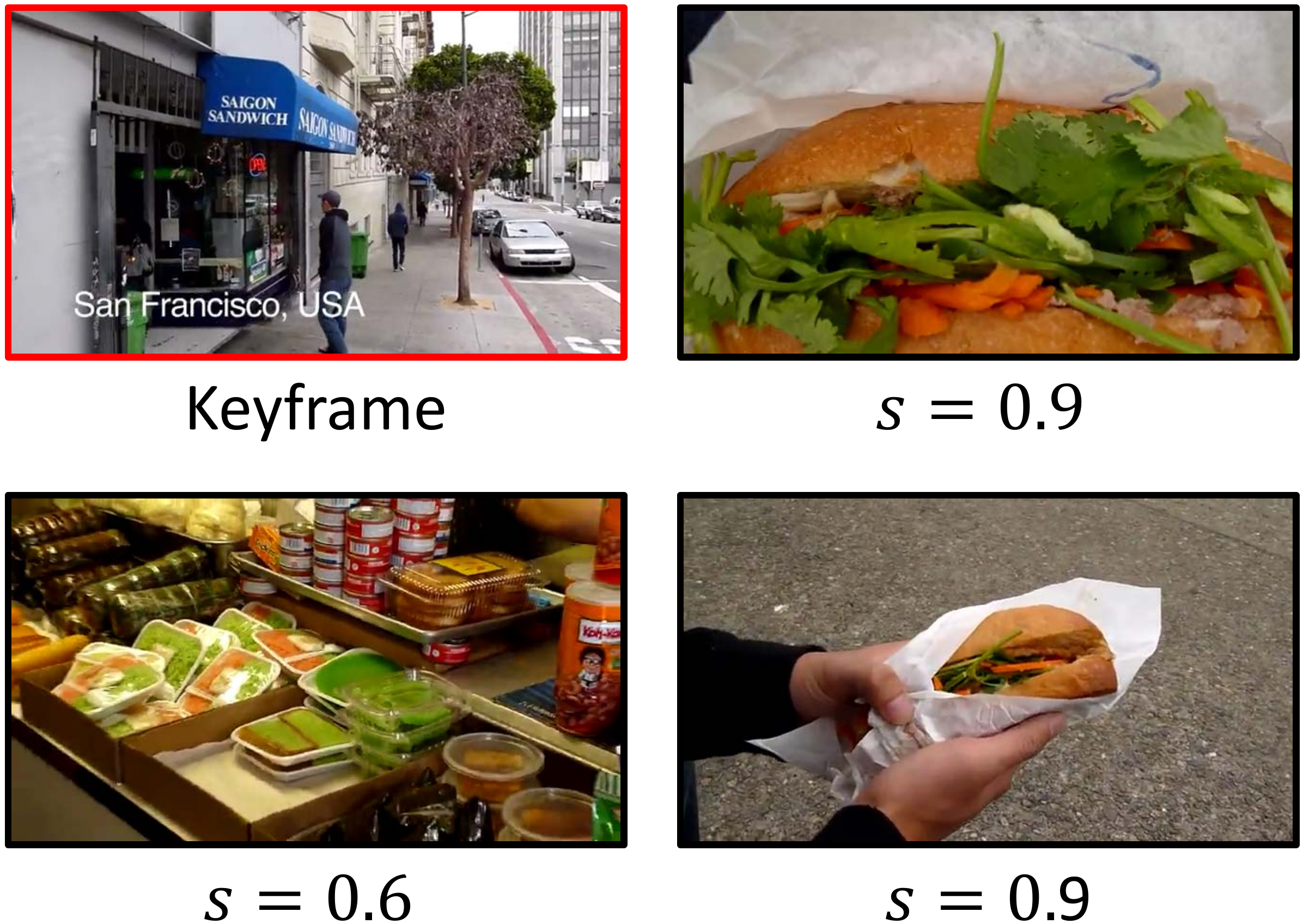}}\hfill
        	\subfigure[(c) Iteration 5]{\includegraphics[width=0.323\linewidth]{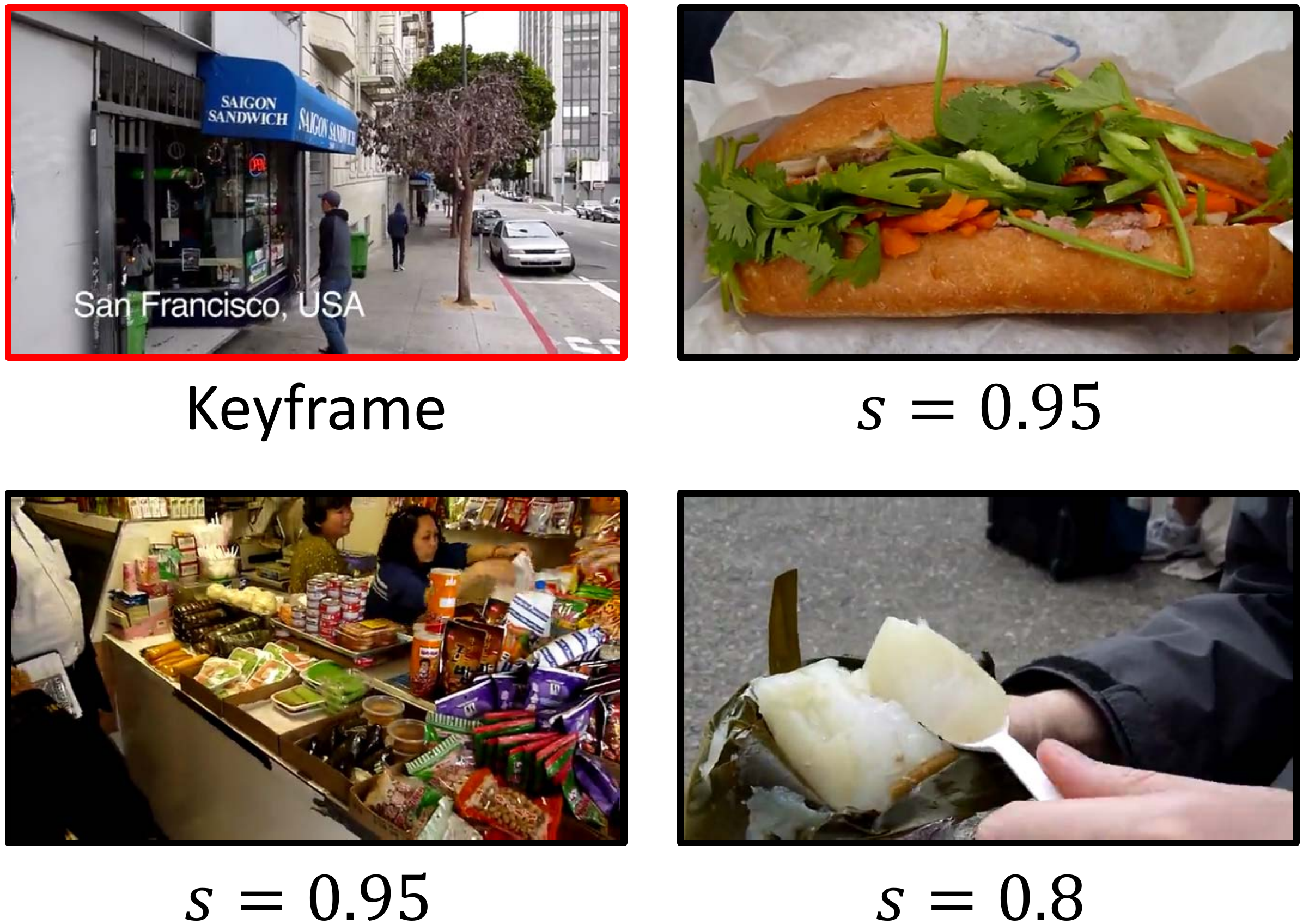}}\hfill\\
        	\caption{
        	Convergence of VideoGraph: We depict frames that have top3 affinity value with a selected keyframe at (a) iteration 1, (b) iteration 3, and (c) iteration 5; where $s$ denotes the averaged and normalized user-annotated importance scores which range from 0 to 1. The keyframes are progressively connected with high affinity as graph refinement repeats in VideoGraph.
        }\label{fig:6}
    \end{figure*}

    \subsection{Recursive Graph Refinement}
    Although the visual representations are guided by the language query, the initial spatial and temporal graphs are highly biased towards visual similarity since the initial object node representations are extracted from the pretrained visual network.
    In addition, constructing graph edges by connecting all objects and frames aggregates redundant and noisy information for video summarization.
    Therefore, unrelated node pairs should be disconnected.
    To this end, the SRR and TRR networks interact with each other by iteratively estimating relative residual adjacency matrices to refine the spatial and temporal graphs.
    Formally, intermediate object node representations for the $t$-th frame $\mathbf{Z}_{t}^{k}$ are derived from \equref{equ:Z_S} such that, $\mathbf{Z}_{t}^{k} = \mathcal{F}(\mathbf{Z}_{t}^{k-1}, \mathbf{S}_{t}^{k-1} | \mathbf{W}_{z})$.
    The averaged object node representations $\mathbf{F}^{k}$, i.e., the frame node representations at the $k$-th iteration, are used to estimate the residual between the previous and current temporal adjacency matrices as
    \begin{equation}
    \begin{split}
        \mathbf{A}^{k} - \mathbf{A}^{k-1} & = d\mathbf{A}^{k} \\
                                         & = \sigma(\frac{(\mathbf{W}_{\theta_{S}}\mathbf{F}^{k})^\top(\mathbf{W}_{\phi_{S}}\mathbf{F}^{k})}{||\mathbf{W}_{\theta_{S}}\mathbf{F}^{k}||_{2} \cdot ||\mathbf{W}_{\phi_{S}}\mathbf{F}^{k}||_{2}}),
    \end{split}
    \end{equation}
    where $\mathbf{W}_{\theta_{S}}$ and $\mathbf{W}_{\phi_{S}}$ are two different weight matrices to be learned to enable computing the affinity in the linear embedding space, and $\sigma(\cdot)$ is an activation function.
    The refined temporal adjacency matrix $\mathbf{A}^{k} = \mathbf{A}^{k-1} + d\mathbf{A}^{k}$ aggregates the object node representations over frames in the TRR network, and the updated object node representations $\hat{\mathbf{Z}}_{t}$ are obtained following \equref{equ:Z_hat}.
    
    Similarly, the spatial adjacency matrix $\mathbf{S}_{t}^{k}$ for the $t$-th frame at the $k$-th iteration is refined from $\hat{\mathbf{Z}}_{t}$ as
    \begin{equation}
    \begin{split}
        \mathbf{S}_{t}^{k} & = \mathbf{S}_{t}^{k-1} + d\mathbf{S}_{t}^{k} \\
                        & = \mathbf{S}_{t}^{k-1} + \sigma(\frac{(\mathbf{W}_{\theta_{T}}\hat{\mathbf{Z}}_{t}^{k-1})^\top(\mathbf{W}_{\phi_{T}}\hat{\mathbf{Z}}_{t}^{k-1})}{||\mathbf{W}_{\theta_{T}}\hat{\mathbf{Z}}_{t}^{k-1}||_{2} \cdot ||\mathbf{W}_{\phi_{T}}\hat{\mathbf{Z}}_{t}^{k-1}||_{2}}),
    \end{split}
    \end{equation}
    where $\mathbf{W}_{\theta_{T}}$ and $\mathbf{W}_{\phi_{T}}$ are weight matrices to be learned.
    
    The final spatial and temporal adjacency matrices are then formulated in a recurrent manner as
    \begin{equation}
        \mathbf{A}^{K} = \mathbf{A} + \sum_{k=1}^{K}{d\mathbf{A}^{k}}, \quad \mathbf{S}_{t}^{K} = \mathbf{S}_{t} + \sum_{k=1}^{K}{d\mathbf{S}_{t}^{k}},
    \end{equation}
    where $K$ denotes the maximum iteration, and $\mathbf{A}$ and $\mathbf{S}_{t}$ are the initial adjacency matrices.
    In contrast to \cite{Wang18graph} where the adjacency matrix is constructed from the input feature representations and fixed, we obtain the refined adjacency matrices through iterative refinement with the output representations.
    Repeatedly inferring the relative residual adjacency matrices facilitates fast and stable convergence for video summarization and gradually incorporates the spatial and temporal relationships.
    Furthermore, residual shortcut connections can transfer information easily (directly propagating information) by addition rather than accumulation by multiplication.
    In practice, the object and frame nodes initially connected by the visual similarity are progressively linked into the subset of nodes based on semantic relationships, as shown in \figref{fig:6}.

    \subsection{Summarization Network}
    Our final goal is to classify each node in the temporal graph.
    As illustrated in \figref{fig:3}, the final object node representations $\hat{\mathbf{Z}}^{K}$ and temporal adjacency matrix $\mathbf{A}^{K}$ are fed into the summarization network to create a video summary.
    We first feed the final object node representations $\hat{\mathbf{Z}}^{K}$ into the average pooling layer to produce final frame node representations $\tilde{\mathbf{F}}^{K}$.
    We then append the graph convolutional layers, followed by an activation function to obtain a binary summary score $\mathbf{Y}$ indicating the probability that each node is a keyframe or background, such that,
    \begin{equation}
        \mathbf{Y} = \mathcal{F}(\tilde{\mathbf{F}}^K, \mathbf{A}^K|\mathbf{W}_{s}),
    \end{equation}
    where $\mathbf{W}_{S}$ is a set of learnable parameters of the summarization network.
    If the probability corresponding to the keyframe is higher than that of the background, the node is contained in the set of keyframes $\mathcal{K}$.

    \subsection{Loss Functions}
    To utilize our model in a practical environment, we verify VideoGraph in supervised and unsupervised settings.
    Specifically, we train VideoGraph with four loss functions for a supervised setting, including the classification, sparsity, reconstruction, and diversity loss functions; and three loss functions (omitting the classification loss) for an unsupervised setting.
    
    For the classification loss $\mathcal{L}_{c}$, we apply a weighted binary cross entropy loss between the output of the summarization network and the groundtruth label to address imbalances between the number of keyframes and background frames, following the previous works~\cite{Rochan18,vs_lstm}:

    \begin{equation}\label{eq:Lcb}
    \mathcal{L}_{c}^{\text{bin}} = -\frac{1}{T} \sum_{t=1}^{T}{w_t[ y^{*}_{t}\text{log}(y_{t})}+(1-y^{*}_{t})\text{log}(1-y_{t})],
    \end{equation}
    where $y^{*}_{t}$ is the groundtruth binary label of the $t$-th frame.
    Each node is weighted by $w_t = \omega_{m} / \omega_{t}$, where $\omega_{t}=\frac{|\mathcal{K}|}{T}$ for keyframes and $\omega_{t}=1-\frac{|\mathcal{K}|}{T}$ for the background frames with the number of keyframes $|\mathcal{K}|$.
    Since there are two classes (keyframe or not), we set the median frequency $\omega_{m}$ to $0.5$.

    In addition to the binary label, the importance score for each frame has been utilized to train the model~\cite{hsa-rnn, rsgn, vjmht, li23}.
    Since the score label represents the relative importance of each frame, it provides more fine information to the model.
    We alternatively apply the mean squared error between the groundtruth importance score $y_t^*$ and the predicted score $y_t$ as the classification loss:
    \begin{equation}\label{eq:Lcs}
        \mathcal{L}_c^{\text{sco}} = \frac{1}{T}\sum_{t=1}^{T} ||y_t - y_t^*||_2^2.
    \end{equation}

    In practice, a long video can be summarized into a sparse subset of keyframes.
    Based on this intuition, our model learns the parameters to construct sparse spatial and temporal graphs for video summarization.
    We employ an entropy function~\cite{entropy} of the spatial and temporal adjacency matrices to enforce this constraint as follows:
    \begin{equation}
    \begin{split}
        & \mathcal{L}_{s}^{s} = -\sum_{t=1}^{T}\sum_{i=1}^{N}\sum_{j=1, j \neq i}^{N} s_{t, ij}^{K}\log{s_{t, ij}^{K}},   \\
        & \mathcal{L}_{s}^{t} = -\sum_{i=1}^{T}\sum_{j=1, j \neq i}^{T} a_{ij}^{K}\log{a_{ij}^{K}},
    \end{split}
    \end{equation}
    where $s_{t, ij}^{K} \in \mathbf{S}_{t}^{K}$ and $a_{ij}^{K} \in \mathbf{A}^{K}$.
    The sparsity loss is then given by the combination of two entropy functions as
    \begin{equation}\label{equ:entropy}
        \mathcal{L}_{s} = \mathcal{L}_{s}^{s} + \rho \cdot \mathcal{L}_{s}^{t},
    \end{equation}
    where $\rho$ controls the balance between the two entropy functions.
    By minimizing uncertain and noisy edge values, the entropy function enables learning more discriminative connections in fully connected graphs than \cite{park20}, which uses $L_1$ normalization as the sparsity loss.

    The keyframes should be visually diverse to ensure the summary represents the entire video story~\cite{vs_lstm, Rochan18}.
    The reconstruction and diversity loss functions are employed to facilitate diverse keyframe selection.
    To reconstruct the original frame node representations $\mathbf{X}_{\mathcal{K}}$, we decode the final frame node representations of the selected keyframes $\hat{\mathbf{F}}_{\mathcal{K}}^{K}$ by applying consecutive $1\times1$ convolutions so that the dimensionality of the decoded representation is the same as the original node representations.
    The decoded and original frame node representations are concatenated and fed into an additional $1\times1$ convolution layer to obtain the final reconstructed frame node representations.
    The reconstruction loss $\mathcal{L}_{r}$ is defined as the mean squared error between the reconstructed and original node representations, such that:
    \begin{equation}
        \mathcal{L}_{r} = \frac{1}{|\mathcal{K}|}\sum_{t\in\mathcal{K}}{||\mathbf{x}_{t} - \hat{\mathbf{x}}_{t}||_{2}^{2}},
    \end{equation}
    where $\hat{\mathbf{x}}$ denotes the final reconstructed features.
    Note that, the original frame node representations are derived by averaging the input object node representations in each frame as $\mathbf{x}_{t} = \frac{1}{N} \sum_{n=1}^{N} {\mathbf{o}_{t}^{n}}$, to reduce the computational complexity.

    To ensure the visual diversity of the selected keyframes, we apply a repelling regularizer~\cite{Zhao17} to the reconstructed frame node representations as the diversity loss:
    \begin{equation}\label{equ:Ldiv}
        \mathcal{L}_{d} = \frac{1}{|\mathcal{K}|(|\mathcal{K}| - 1)} \sum_{i \in \mathcal{K}}\sum_{j \in \mathcal{K}, j \neq i} \frac{\hat{\mathbf{x}}_{i}^{T}\hat{\mathbf{x}}_{j}}{||\hat{\mathbf{x}}_{i}||_{2}^{2}\cdot||\hat{\mathbf{x}}_{j}||_{2}^{2}}.
    \end{equation}
    Since the reconstruction and diversity loss functions complement each other, they are applied simultaneously to learn the model.
    
     The final loss function for supervised learning is then,
    \begin{equation}\label{equ:Lsup}
        \mathcal{L}_{sup} = \mathcal{L}_{c}^{*} + \alpha \cdot \mathcal{L}_{s} + \beta \cdot \mathcal{L}_{d} + \gamma \cdot \mathcal{L}_{r},
    \end{equation}
    where $\mathcal{L}_{c}^{*}$ is the classification loss according to the groundtruth types (i.e., binary or importance score), $\alpha$, $\beta$, and $\gamma$ control the trade-off between the four loss functions.
    The graph modeling scheme in VideoGraph has also been extended for unsupervised video summarization with a simple modification to the loss functions.
    Since the groundtruth summary cannot be used as supervision in an unsupervised manner, the final loss function for unsupervised learning is represented as
    \begin{equation}\label{equ:Lunsup}
        \mathcal{L}_{unsup} = \alpha \cdot \mathcal{L}_{s} + \beta \cdot \mathcal{L}_{d} + \gamma \cdot \mathcal{L}_{r},
    \end{equation}
    where $\alpha$, $\beta$, and $\gamma$ are balancing parameters to control the trade-off between the three terms. 

\section{Experiments}\label{sec:exp}

    \subsection{Experimental Settings}
    
    \subsubsection{Datasets}
    We evaluated our model on two standard video summarization datasets (i.e., SumMe~\cite{Gygli14} and TVSum~\cite{Song15}), and on the egocentric video dataset (UT Egocentric~\cite{lee12}).
    The SumMe dataset consists of 25 videos capturing multiple events, including cooking and sports, and the lengths of the videos vary from 1.5 to 6.5 minutes.
    The TVSum dataset contains 10 categories from the MED task, including 5 videos per category sampled from YouTube.
    The contents of the videos are diverse, similar to the SumMe, and the video lengths vary from 1 to 5 minutes.
    These datasets provide frame-level importance scores annotated by several users.
    We used additional datasets, the YouTube~\cite{de11} and OVP~\cite{OVP} datasets, to augment the training data.
    The UT egocentric~\cite{lee12} dataset consists of four videos recorded by wearable first-person view cameras.
    Each video contains 3-5 hours long multiple diverse events captured in uncontrolled environments.
    The QFVS~\cite{qfvs} dataset provides generic summaries and query-focused summaries for each video in the UT egocentric dataset.
    For the generic summaries, each video is divided into multiple clips and annotated by selecting clips for the generic contents of each video by three users and averaging all the annotations.
    For the query-focused summaries, three users selected clips corresponding to given language queries (e.g. book and street).
    We use the groundtruth summaries to evaluate the generic and query-focused summarization performance.
    
    \begin{table}[t]
    \centering
    \caption{The format of produced groundtruth (GT) and converted training and testing GT for different datasets. In training, we convert from multiple user frame-level annotations to a single set of the keyframes for the SumMe~\cite{Gygli14} and TVSum~\cite{Song15} datasets. In testing, frame-level scores are converted to the interval-based keyshot annotations for the TVSum dataset following~\cite{Zhang16CVPR,vs_lstm}. Note that OVP and YouTube datasets are not used in the testing phase.}
    \begin{tabular}{lccc}
			\toprule
			Dataset  & \makecell{Produced \\GT} & \makecell{Training \\GT} & \makecell{Testing \\GT}\tabularnewline
			\midrule
			SumMe  & frame scores & keyframes & keyshots \tabularnewline
			TVSum  & frame scores & keyframes & keyshots    \tabularnewline
			OVP  & keyframes & keyframes & -  \tabularnewline
			YouTube  & keyframes & keyframes & -  \tabularnewline
			\bottomrule
        \end{tabular}\label{tab:groundtruth}
    \end{table}

    \subsubsection{Groundtruth}
    The groundtruth annotations of the datasets used in the experiments (i.e., SumMe, TVSum, OVP, and YouTube) are provided in various formats, as shown in \tabref{tab:groundtruth}.
    Following previous works~\cite{gong14,vs_lstm,Rochan18}, we generated a single set of keyframes for each training video from multiple user annotations.
    Specifically, we labeled the frames selected for summary as 1 and others as 0.
    We converted the predicted keyframes and groundtruth keyframes to the interval-based keyshot summaries for the testing videos used in the performance evaluations.
    Since the TVSum dataset only provides keyframe annotations, we convert the keyframe annotations to keyshot-based summaries following the steps in~\cite{vs_lstm}: 
    1) apply kernel temporal segmentation (KTS)~\cite{potapov14} to generate temporal segments in the form of disjoint intervals; 
    2) compute the average importance score for each segment and assign the score to the frames in each interval; and
    3) apply the knapsack algorithm \cite{Song15} to select frames such that the length of the keyshot groundtruth is not exceeded a certain threshold.
    Consequently, the keyshot-based annotations are represented by a binary vector (i.e., 1 for keyframes and 0 for the background) where the number of elements is equal to the number of frames in the video.

    \subsubsection{Data configuration}
    \noindent\textbf{Generic video summarization.}
    We evaluate our model on the SumMe and TVSum datasets with three different data configurations for generic video summarization: standard, augmented, and transfer data settings.
    The standard data setting selects training and testing videos from the same dataset by splitting $80\%$ of the videos for training and the remainder for testing.
    The augmented data setting uses the other three datasets to augment training data.
    For example, $80\%$ videos of the SumMe and whole videos of the TVSum, OVP, and YouTube datasets are used for training, and $20\%$ videos of the SumMe are used to evaluate the performance on SumMe.
    Not surprisingly, augmenting the training data derives improved performances similar to recent works~\cite{Zhang18,Rochan18}.
    The more challenging setting, the transfer data setting introduced in \cite{vs_lstm}, uses other available datasets to train the model, and the target dataset is only used to evaluate the performance.
    In addition, we evaluate the model on the QFVS dataset using the generic summary groundtruth.
    We run four rounds of experiments to evaluate the generic video summarization performance by splitting four videos into 2, 1, and 1 for training, validation, and testing, respectively, following prior works~\cite{qfvs,clipit}.
    We train the model with three different levels of the language query: using video data only (w/Video only), the word-level language query (w/Word-level query), and the sentence-level language query (w/Sentence-level query).

    \noindent\textbf{Query-focused video summarization.}
    We provide the query-focused video summarization performance evaluated on the QFVS dataset.
    Each video in the dataset is divided into 45 different language queries (e.g. book\_chair, car\_market), and the groundtruth corresponding to each query is provided.
    We use two videos for training, one video for validation, and the remaining one video for evaluation for query-focused video summarization.
    Different from generic video summarization, we train the model with captions provided by the dataset in a supervised manner.

    \begin{table}[!t]
	\centering
	\caption{Network configuration of VideoGraph. `Lan-emb.' denote linear layers to embed the sentence-level queries and `Lan-emb.$^*$' is for the word-level queries. `MHCA-emb.' is a linear layer in the MHCA, `G-conv$k$s' and `G-conv$k$t' denote the graph convolutional layer in the SRR and TRR networks, respectively. `Common Layers' are used in the feature extraction network and MHCA.}
		\begin{tabular}{
			>{\raggedright}m{0.22\linewidth} |
			>{\centering}m{0.18\linewidth} >{\centering}m{0.1\linewidth}
			>{\centering}m{0.2\linewidth}}
                \toprule
                \multicolumn{4}{c}{Common Layers}   \tabularnewline
                \midrule
                Layer & Ch I/O & BN & Activation \tabularnewline
			\midrule
                Obj-emb.   &   2048/1024   &   -    &   -  \tabularnewline
                Lan-emb.    &   2048/1024   &   -    &   -  \tabularnewline
                Lan-emb.$^*$   &   128$\times W$/1024   &   -    &   -  \tabularnewline
                MHCA-emb.    &   1024/512    &   -   &   -   \tabularnewline
                \midrule
			\multicolumn{4}{c}{Graph Layer 1} \tabularnewline
			\midrule
			Layer & Ch I/O & BN & Activation \tabularnewline
			\midrule
			G-conv1s & 512/256 & \cmark & ELU($\cdot$) \tabularnewline
			G-conv1t & 256/512 & \cmark & ELU($\cdot$) \tabularnewline
			\midrule
			\multicolumn{4}{c}{Graph Layer 2} \tabularnewline
			\midrule
			Layer & Ch I/O & BN & Activation \tabularnewline
			\midrule
			G-conv1s    &   512/256    & \cmark & ELU($\cdot$)     \tabularnewline
			G-conv2s    &   256/256     & \cmark & ELU($\cdot$)     \tabularnewline
			G-conv1t    &   256/256     & \cmark & ELU($\cdot$)     \tabularnewline
			G-conv2t    &   256/512    & \cmark & ELU($\cdot$)    \tabularnewline
			\midrule
			\multicolumn{4}{c}{Graph Layer 3} \tabularnewline
			\midrule
			Layer & Ch I/O & BN & Activation \tabularnewline
			\midrule
			G-conv1s    &   512/256    &  \cmark & ELU($\cdot$)  \tabularnewline
			G-conv2s    &   256/256     &  \cmark & ELU($\cdot$) \tabularnewline
			G-conv3s    &   256/256     &  \cmark & ELU($\cdot$) \tabularnewline
			G-conv1t    &   256/256     &   \cmark & ELU($\cdot$) \tabularnewline
			G-conv2t    &   256/256    &    \cmark & ELU($\cdot$)  \tabularnewline
			G-conv3t    &   256/512    &    \cmark & ELU($\cdot$)  \tabularnewline
			\midrule
			\multicolumn{4}{c}{Graph Layer 4} \tabularnewline
			\midrule
			Layer  & Ch I/O & BN & Activation \tabularnewline
			\midrule
			G-conv1s    &   512/256    &  \cmark & ELU($\cdot$)  \tabularnewline
			G-conv2s    &   256/256     &  \cmark & ELU($\cdot$) \tabularnewline
			G-conv3s    &   256/256     &  \cmark & ELU($\cdot$) \tabularnewline
			G-conv4s    &   128/128     &  \cmark & ELU($\cdot$) \tabularnewline 
			G-conv1t    &   128/128     &   \cmark & ELU($\cdot$) \tabularnewline
			G-conv2t    &   128/256    &    \cmark & ELU($\cdot$)  \tabularnewline
			G-conv3t    &   256/256    &     \cmark & ELU($\cdot$)  \tabularnewline
			G-conv4t    &   256/512    &    \cmark & ELU($\cdot$)  \tabularnewline
			\midrule
			
			\multicolumn{4}{c}{Summarization Layers} \tabularnewline
			\midrule
			Layer & Ch I/O & BN & Activation \tabularnewline
			\midrule
			Conv-sum1   &   512/256    &   \cmark  & ReLU($\cdot$)  \tabularnewline
			Conv-sum2   &   256/2       &   \cmark  & Softmax($\cdot$)   \tabularnewline
			Conv-recon1 &   512/2048   &   \cmark  & ELU($\cdot$)    \tabularnewline
			Conv-recon2 &   4096/2048   &   -    &   -    \tabularnewline
			\bottomrule
		\end{tabular}
		\label{tab:configuration}	
    \end{table}

    \subsubsection{Evaluation metrics}
    We evaluated our method using the keyshot-based metrics commonly used in recent works~\cite{Rochan18,Rochan19}.
    Let $\mathbf{Y}$ and $\mathbf{Y}^{*}$ be the predicted keyshot summary and groundtruth summary created by multiple users, respectively.
    The precision ($P$) and recall ($R$) are defined as
    \begin{equation}
    \begin{split}
        P = \frac{\text{overlap between $\mathbf{Y}$ and $\mathbf{Y}^{*}$}}{\text{total duration of $\mathbf{Y}$}}, \\
        R = \frac{\text{overlap between $\mathbf{Y}$ and $\mathbf{Y}^{*}$}}{\text{total duration of $\mathbf{Y}^{*}$}}.
    \end{split}
    \end{equation}
    We compute the F-score to measure the quality of the summary with precision and recall:
    \begin{equation}
        F\text{-}score = \frac{2 \times {P} \times {R}}{P + R}.
    \end{equation}
    For datasets with multiple groundtruth summaries, we follow the standard approaches~\cite{vs_lstm,Gygli14,Gygli15} to calculate the metrics for the videos.
    
    We also evaluated our method using rank-based metrics, Kendall's $\tau$~\cite{Kendall} and Spearman's $\rho$~\cite{Spearman} correlation coefficients, following \cite{Otani19}.
    To compute the correlation coefficients, we first rank the video frames according to their probability of being a keyframe and the annotated importance scores.
    After that, we compare the generated ranking with each annotated ranking.
    The correlation scores are then computed by averaging the individual results.
    
\subsection{Implementation Details}\label{sec:expset}
    \subsubsection{Network Architecture}\label{sec:network}
    The VideoGraph is composed of four main parts: a \textit{feature extraction network} to extract the initial object node representations and the language query representation from the pretrained feature extractors; \textit{spatial and temporal relation reasoning networks} that construct language-guided spatial and temporal graphs, propagate the object and frame nodes along spatial and temporal edges, and refine the spatial and temporal graphs by recursively estimating relative adjacency matrices; and a \textit{summarization network} to predict keyframes.
    Although we use three graph convolution layers for the spatial and temporal graphs in each relation reasoning network, we can use fewer or more graph convolution layers.
    The overall configuration of VideoGraph corresponding to the number of graph convolution layers is shown in \tabref{tab:configuration}.
     
    \subsubsection{Feature extraction}
    \noindent\textbf{Visual feature extraction.}
    To train VideoGraph, we uniformly sampled frames for every video at 2 frames per second, as described in \cite{vs_lstm}.
    We fed the sampled frames into Faster R-CNN~\cite{faster-rcnn} that uses ResNet-101~\cite{he16} as the backbone with bottom-up attention~\cite{bottom-up} pretrained on Visual Genome~\cite{visual-genomes}.
    The objects in each frame are detected by the model and applied non-maximum suppression~\cite{nms} for each class with an intersection over union (IoU) threshold of 0.7.
    We select the top $N$ region of interests (RoIs) with the highest detection confidence scores and extract $2048$-dimensional activations as an object node feature representation $\mathbf{o}_{t}^{n}$ for each selected RoI after average pooling.
    Although the maximum number of objects $N$ in each frame is set to $36$, we evaluated the model primarily using the top $16$ object feature representations based on the ablation study.
    Note that the object detector is frozen while training.
    
    \noindent\textbf{Language feature extraction.}
    In generic video summarization, we generate two types of language queries, i.e., sentence-level queries and word-level queries.
    To obtain the sentence-level language queries, we use the bi-modal transformer dense video captioning model~\cite{captionmodel}, which generates a multiple sentence description, following~\cite{clipit} for fair comparisons.
    We first embed each sentence of the description using the CLIP~\cite{clip} text encoder and fuse them using the MLP layer.
    For the word-level language queries, we select $W$ words from the detected object classes based on the detected frequency.
    We set $W$ differently according to the dataset through the ablation study.
    Each word is embedded by the CLIP text encoder and the MLP layer and is fused using an additional MLP layer.
    In the case of the user-defined queries, we identically use the CLIP text encoder for each query and fuse all query embeddings using the MLP layer.
    As the same with the visual feature 

    \clearpage
    \begin{sidewaystable*}[hbt!]
    \centering
    \small
    \caption{Comparison with state-of-the-art methods on the SumMe~\cite{Gygli14} and TVSum~\cite{Song15} datasets, with various data configurations including standard data, augmented data, and transfer data settings. ${^\dagger}$ used partial supervision. We denote the usage of language query on the column of `Language'.}
    \scalebox{0.9}{\begin{tabular}{llccccccc}\toprule
    \multirow{2}{*}[-0.3em]{Supervision}   &   \multirow{2}{*}[-0.3em]{Method}  &   \multirow{2}{*}[-0.3em]{Language} & \multicolumn{3}{c}{SumMe}     & \multicolumn{3}{c}{TVSum}     \\ \cmidrule(lr){4-6} \cmidrule(lr){7-9} 
            &   &   & Standard & Augment & Transfer & Standard & Augment & Transfer \\ \midrule
        \multirow{11}{*}{Unsupervised}    &   Mahasseni  et al.\cite{Mahasseni17}    &   \xmark   & 39.1 &  43.4    &   -    & 51.7  &  59.5    & -   \tabularnewline
        &   Rochan  et al.\cite{Rochan18}(SUM-FCN$_{unsup}$)   &   \xmark   &  41.5  &  -   & 39.5    &   52.7   &    -     &  -\tabularnewline
        &   Rochan  et al.$^{\dagger}$\cite{Rochan19} &   \xmark   &  -   & 48.0  & 41.6 & -  &   56.1  & 55.7\tabularnewline
        &   He  et al.\cite{He19}    &   \xmark   &   46.0   &   47.0    &   44.5   &   58.5   &   58.9    &   57.8   \tabularnewline
        &   Yuan  et al.\cite{Yuan19} &   \xmark   &  41.9 & - & - & 57.6    &   -    &   -\tabularnewline
        &   Park  et al.\cite{park20} &   \xmark   &  49.8  & 52.1 &   47.0   &  59.3   &   61.2   &   57.6  \tabularnewline 
        &   Wu  et al.\cite{era} &   \xmark   &  23.2  & - &   -   &  58.0   &   -   &   -  \tabularnewline 
        &   Li et al.\cite{vjmht} &   \xmark   &  50.6  & 51.7 &   46.4   &  60.9   &   61.9   &   58.9        \tabularnewline 
        &   Narasimhan  et al.\cite{clipit} &   \cmark   &  52.5  & 54.7 &   50.0   &  63.0   &   65.7   &   62.8  \tabularnewline 
        \cmidrule{2-9}
        &   \textbf{VideoGraph} w/Video only &   \xmark   &  {51.2}  & {54.0} &   {48.5}   &  {61.1}   &   {62.3}   &   {58.7}   \tabularnewline
        &   \textbf{VideoGraph} w/Word-level query  &   \cmark   &  \textbf{55.0}   &   \textbf{56.9} &  \textbf{53.0}   & \textbf{66.8}   & \textbf{68.1}   & \textbf{65.5}
        \tabularnewline 
        &   \textbf{VideoGraph} w/Sentence-level query  &   \cmark   &  {53.4}   &   {55.3} &  {52.9}   & {65.6}   & {67.2}   & {64.7}     \tabularnewline   
        \midrule
        \multirow{19}{*}{Supervised}    &   Zhang  et al.~\cite{Zhang16CVPR}    &   \xmark   & 40.9    &   41.3    &  38.5    &  - &   -   &   -\tabularnewline
        &   Zhang  et al.\cite{vs_lstm}   &   \xmark   & 38.6   &   42.9    &   41.8    &   54.7    &   59.6    &   58.7\tabularnewline
        &   Mahasseni  et al.\cite{Mahasseni17} (SUM-GAN$_{sup}$)   &   \xmark   & 41.7 &  43.6    &   -    & 56.3  &  61.2    & -   \tabularnewline
        &   Zhao  et al.\cite{Zhao17} &   \xmark   & 44.3 & - & - & 62.1 & - & - \tabularnewline
        &   Rochan  et al.\cite{Rochan18}(SUM-FCN)   &   \xmark   &  47.5  &  51.1   & 44.1    &   56.8   &   59.2     &   58.2\tabularnewline
        &   Rochan  et al.\cite{Rochan18}(SUM-DeepLab) &   \xmark   & 48.8   &    50.2   & 45.0      & 58.4  & 59.1 &  57.4 \tabularnewline
        &   Zhou  et al.\cite{Zhou18} &   \xmark   & 42.1 & 43.9  & 42.6  & 58.1  & 59.8  & 58.9 \tabularnewline
        &   Zhang  et al.\cite{Zhang18} &   \xmark   & 44.9 & - & - & 63.9 & - & -\tabularnewline
        &   He  et al.\cite{He19}    &   \xmark   &   47.2   &   -    &   -   &   59.4   &   -    &   -   \tabularnewline
        &   Ji  et al.\cite{avs} &   \xmark   &  44.4 &  46.1 &  -    & 61.0   &    61.8   &  - \tabularnewline
        &   Zhao  et al.\cite{hsa-rnn} &   \xmark   &  42.5 &  - &  -    & 58.4   &    -   &  - \tabularnewline
        &   Park  et al.\cite{park20} &   \xmark   &  51.4 &  52.9 &  48.7    & 63.9   &    65.8   &  60.5 \tabularnewline
        &   Ji  et al.\cite{adsum} &   \xmark   &  46.1 &  47.6 &  -    & 64.5   &    65.8   &  - \tabularnewline
        &   Zhu  et al.\cite{dsnet} &   \xmark   &  51.2  & 53.3 &   47.6   &  61.9   &   62.2   &   58.0  \tabularnewline
        &   Zhao  et al.\cite{rsgn} &   \xmark   &  45.0  & 45.7 &   44.0   &  60.1   &   61.1   &   60.0  \tabularnewline
        &   Jiang  et al.\cite{iptnet} &   \xmark   &  54.5  & 56.9 &   49.2   &  63.4   &   64.2   &   59.8  \tabularnewline
        &   Narasimhan  et al.\cite{clipit} &   \cmark   &  54.2  & 56.4 &   51.9   &  66.3   &   69.0   &   65.5  \tabularnewline 
        &   Li  et al.\cite{li23} &   \cmark   &  50.7  & - &   -   &  60.4   &   -   &   -  \tabularnewline 
        \cmidrule{2-9}
        &   \textbf{VideoGraph} w/Video only &   \xmark   &  {54.7}   &   {55.6} &  {55.4}   & {65.2}   & {67.3}   & {61.9}     \tabularnewline 
        &   \textbf{VideoGraph} w/Word-level query &   \cmark   &  \textbf{58.1}   &   \textbf{59.4} &  \textbf{55.6}   & \textbf{69.5}   & \textbf{71.4}   & \textbf{67.7}     \tabularnewline 
        &   \textbf{VideoGraph} w/Sentence-level query  &   \cmark   &  57.0   &   {58.6} &  {54.3}   & {68.4}   & {70.7}   & {67.1}     \tabularnewline 
        \bottomrule
    \end{tabular}}
    \label{tab:1}
    \end{sidewaystable*}
    \clearpage
    \newpage

    \noindent extraction, the caption generator and pretrained CLIP encoder are frozen while training.

    \begin{table}[htbp!]
    \small
    \centering
    \caption{
    Kendall's $\tau$~\cite{Kendall} and Spearman's $\rho$~\cite{Spearman} correlation coefficients computed by VideoGraph with the word-level query and state-of-the-art approaches on the TVSum benchmark~\cite{Song15}. We report the types of the groundtruth for training, i.e., binary (B) and frame score (S) labels.
    }
    \begin{tabular}{l ccc}
        \toprule
        Method &    Label   & $\tau$ &  $\rho$ \tabularnewline
        \midrule
        Human & -   & 0.177 & 0.204 \tabularnewline
        Groundtruth &   S   &   0.364   &   0.456  \tabularnewline \midrule
        Zhang et al.~\cite{vs_lstm} &   B  & 0.042 & 0.055 \tabularnewline
        Zhou et al.~\cite{Zhou18}   &   -   & 0.020 &  0.026 \tabularnewline
        Park et al.~\cite{park20} &   B  & 0.094 & 0.138 \tabularnewline
        Narasimhan et al.~\cite{clipit} &   B  & 0.108 & 0.147 \tabularnewline
        Zhao et al.~\cite{hsa-rnn} &   S  & 0.082 &  0.088 \tabularnewline
        Zhou et al.~\cite{rsgn} &   S  & 0.083 &  0.090 \tabularnewline
        Li et al.~\cite{vjmht} &   S  & 0.097 &  0.105 \tabularnewline
        Li et al.~\cite{li23}   &   S   & 0.181 &  0.238 \tabularnewline
        \midrule
        \textbf{VideoGraph} &   B  & 0.162 & 0.188 \tabularnewline
        \textbf{VideoGraph}&   S  & $\mathbf{0.204}$ & $\mathbf{0.242}$\tabularnewline
        \bottomrule
        \end{tabular}
    \label{tab:coefficient}
    \end{table}
    
    \subsubsection{Training and optimization}
    Although VideoGraph can be trained independently of the lengths of the videos, we uniformly sampled frames with fixed temporal length $T=320$ from each training video for TVSum and SumMe to learn the network parameters with a mini-batch size of 5.
    We train VideoGraph for 200 epochs with the Adam optimizer~\cite{Adam}.
    The learning rate is set to $10^{-4}$ and decayed by a factor of $0.1$ every 80 epochs.
    We use four heads each in the MHCA of the SRR and TRR networks, the projected query, key, and value have 256 dimensions.
    The scaling factors $\lambda_o$ and $\lambda_f$ in \equref{equ:S} and \equref{equ:A} are set to $1.6$ and $30$, respectively.
    We set $\rho = 5$ in \equref{equ:entropy}, $\alpha = 0.0001$, and $\beta = \gamma = 0.1$ in \equref{equ:Lsup} for supervised learning, and $\alpha = 0.001$ and $\beta = \gamma = 10$ in \equref{equ:Lunsup} for unsupervised learning.
    The maximum number of iterations $K$ is fixed at five throughout the ablation study (see \figref{fig:8}).
    The quantitative performance for the experiments on TVSum and SumMe is evaluated five times on five random splits of the datasets and reported as the average values.
    
    \subsection{Quantitative Results}\label{sec:mainresults}
    \subsubsection{Evaluation on SumMe and TVSum datasets}
    We train the model for generic video summarization with three different levels of language queries, no language query, word-level language queries, and sentence-level language queries.
    In \tabref{tab:1}, we show our results in comparison to supervised methods (\cite{Zhang16CVPR,vs_lstm,Mahasseni17,Zhao17,Zhang18,Rochan18,Rochan19,He19,avs,park20,adsum,dsnet,rsgn,iptnet,clipit}) and unsupervised video summarization (\cite{Song15,Mahasseni17,Rochan18,Yuan19,Rochan19,He19,park20,era,clipit}) on both SumMe~\cite{Gygli14} and TVSum~\cite{Song15} datasets.

    The comparison between the approaches that not use any language queries shows that our model achieves state-of-the-art performance of $54.7\%$ on SumMe and $65.2\%$ on TVSum in the standard and transfer settings.
    While the performance of \cite{iptnet} in a supervised manner with the augment setting is slightly better than our method, we outperform \cite{iptnet} in other settings (i.e., the standard and transfer settings).
    Especially, the result in the transfer setting proves the generalizability of the proposed method, showing $6.2\%$ and $2.1\%$ performance improvements on the SumMe and TVSum datasets, respectively.
    Our VideoGraph reduces the gap between the supervised and unsupervised approaches.
    It further improves the performance even without using the unpaired data used as additional information for the summary video in \cite{Rochan19}.
    
    The results of the model with language queries demonstrate that the language-guided approach consistently further improves the performance regardless of the types of queries.
    In addition, the comparisons with the state-of-the-art language-guided approach~\cite{clipit} validate the effectiveness of the language-guided graph modeling, showing significant performance improvements with large margins in supervised and unsupervised manners.
    We observe that the model trained with the word-level query, surprisingly, outperforms the model with the sentence-level query on both datasets despite the simplicity of the word-level query.
    We insist on this due to a characteristic of the datasets:
    the SumMe and TVSum datasets consist of short videos that contain no complicated themes without multiple events so that the simple keywords can more suitably represent the whole video than the multiple complicated sentences.
    
    \begin{table*}[!t]
    \small
        \centering
        \caption{Comparison with state-of-the-art methods on the QFVS~\cite{qfvs} dataset for generic video summarization in terms of F-score.}
        \begin{tabular}{llcccc c}\toprule
        Supervision  &   Method  &   Vid 1   &   Vid 2   &   Vid 3   &   Vid 4   &   Avg.    \\  \midrule
        \multirow{6}{*}{Unsupervised}
        &   Zhao et al.~\cite{zhao14}   &   53.06  &   53.80   &   49.91   &   22.31   &   44.77   \tabularnewline    
        &   Narasimhan et al.~\cite{clipit} (Gen. Caption)   &   67.02  &   59.48   &   66.43   &   44.19   &   59.28   \tabularnewline
        &   Narasimhan et al.~\cite{clipit} (GT Caption)    &   73.90   &   66.83   &   75.44   &   52.31   &   67.12   \tabularnewline \cmidrule{2-7}
        &   \textbf{VideoGraph} w/Word-level query     &   69.03  &   62.11   &   68.93   &   48.60   &   62.17   \tabularnewline
        &   \textbf{VideoGraph} w/Sentence-level query     &   71.94  &   65.08   &   71.25   &   50.71   &   64.75   \tabularnewline
        &   \textbf{VideoGraph} w/GT. Caption     &   \textbf{76.34}  &   \textbf{69.52}   &   \textbf{76.86}   &   \textbf{55.29}   &   \textbf{69.50}   \tabularnewline   \midrule
        \multirow{7}{*}{Supervised}
        &   Gygli et al.~\cite{Gygli15}   &   49.51  &   51.03   &   64.52   &   35.82   &   50.22   \tabularnewline
        &   Sharghi et al.~\cite{qfvs}   &   62.66  &   46.11   &   58.85   &   33.50   &   50.29   \tabularnewline
        &   Narasimhan et al.~\cite{clipit} (Gen. Caption)   &   74.13  &   63.44   &   75.86   &   50.23   &   65.92   \tabularnewline
        &   Narasimhan et al.~\cite{clipit} (GT Caption)   &   84.98  &   71.26   &   82.55   &   61.46   &   75.06   \tabularnewline   \cmidrule{2-7}
        &   \textbf{VideoGraph} w/Word-level query     &   76.78  &   67.18   &   77.24   &   53.80   &   68.75   \tabularnewline
        &   \textbf{VideoGraph} w/Sentence-level query     &   79.55  &   69.86   &   78.30   &   57.43   &   71.29   \tabularnewline
        &   \textbf{VideoGraph} w/GT. Caption     &   \textbf{87.02}  &   \textbf{74.62}   &   \textbf{85.47}   &   \textbf{64.19}   &   \textbf{77.83}   \tabularnewline
        \bottomrule
        \end{tabular}
        \label{tab:generic}
    \end{table*}

    \begin{table*}[htb!]
    \small
        \centering
        \caption{Comparison with state-of-the-art methods on the QFVS~\cite{qfvs} dataset for query-focused video summarization in terms of F-score.}
        \begin{tabular}{lccccc}\toprule
           Method  &   Vid 1   &   Vid 2   &   Vid 3   &   Vid 4   &   Avg.    \\  \midrule
           Gong et al.~\cite{gong14}   &   36.59  &   43.67   &   25.26   &   18.15   &   30.92   \tabularnewline
           Sharghi et al.~\cite{sharghi16}   &   35.67  &   42.74   &   36.51   &   18.62   &   33.38   \tabularnewline
           Sharghi et al.~\cite{qfvs}   &   48.68  &   41.66   &   56.47   &   29.96   &   44.19   \tabularnewline
           Narasimhan et al.~\cite{clipit}   &   57.13  &   53.60   &   66.08   &   41.41   &   54.55   \tabularnewline   \midrule
           \textbf{VideoGraph}    &   \textbf{60.21}  &   \textbf{57.50}   &   \textbf{68.71}   &   \textbf{45.38}   &   \textbf{57.95}   \tabularnewline
        \bottomrule
        \end{tabular}
        \label{tab:query-focused}
    \end{table*}

    \tabref{tab:coefficient} compares the correlation coefficients of the proposed VideoGraph, deepLSTM~\cite{vs_lstm}, DR-DSN~\cite{Zhou18}, humans reported in \cite{Otani19}, SumGraph~\cite{park20}, HSA-RNN~\cite{hsa-rnn}, RSGN~\cite{rsgn}, VJMHT~\cite{vjmht}, and SSPVS~\cite{li23}.
    To fairly compare the performance, we report the performance corresponding to the groundtruth types, i.e., binary labels with \eqref{eq:Lcb} and score labels with \eqref{eq:Lcs}, respectively.
    VideoGraph trained with the binary labels outperforms the several baselines (except for \cite{li23}) by large margins despite the coarsely annotated labels.
    The importance score labels significantly improve the performance of VideoGraph, outperforming the previous best performance~\cite{li23} by 0.023 for Kendall's $\tau$ and 0.004 for Spearman's $\rho$.
    
    \subsubsection{Evaluation on QFVS dataset}
    We evaluate the performance on the QFVS~\cite{qfvs} dataset for generic and query-focused video summarization.
    In \tabref{tab:generic}, we report the generic video summarization performance and compare our model with state-of-the-art supervised~\cite{Gygli15,sharghi16,clipit} and unsupervised~\cite{zhao14,clipit} approaches.
    The results demonstrate that our VideoGraph consistently outperforms the prior works in supervised and unsupervised manners.
    Specifically, our model with a sentence-level query can be fairly compared with \cite{clipit} trained with the generated captions (Gen. Caption).
    The proposed method outperforms \cite{clipit} by $5.37\%$ and $5.47\%$ in the average performances in supervised and unsupervised settings, respectively.
    The comparisons between the performance of our model trained with word-level and sentence-level queries validate the effectiveness of sentence-level queries for generic summaries of long videos.
    Since videos in the QFVS dataset contain multiple events with long running times, describing the videos using multiple captions is more effective than using the set of keywords.
    With the groundtruth captions obtained from VideoSet~\cite{videoset}, our model achieves state-of-the-art performances throughout all videos.

    \tabref{tab:query-focused} presents the results of query-focused video summarization for baselines~\cite{gong14,sharghi16,qfvs,clipit} and ours.
    Note that we train the model only in a supervised manner with the groundtruth queries provided from the dataset.
    Similar to generic video summarization, our model shows superior performance, outperforming the strongest baseline~\cite{clipit} by $3.4\%$ on the average performance.

    \begin{figure}[!t]
    	\centering
        	\renewcommand{\thesubfigure}{}
        	\subfigure[(a) SumGraph~\cite{park20} (F-Score: 50.3)]{\includegraphics[width=1\linewidth]{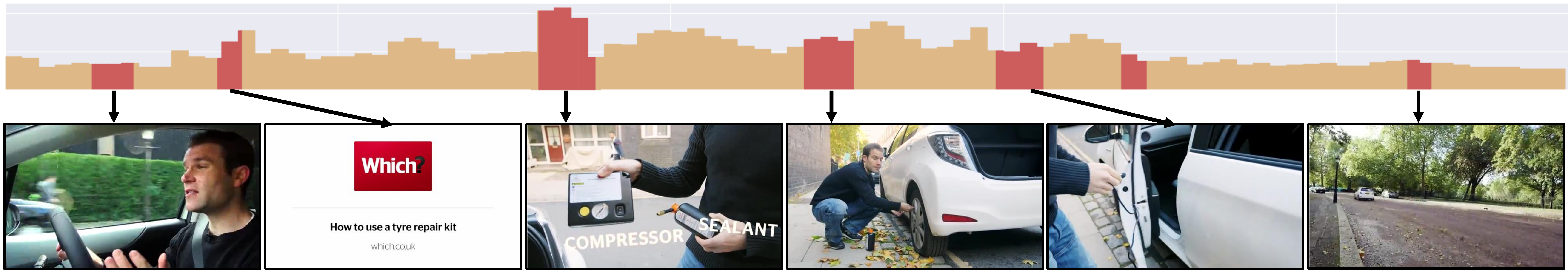}}\hfill\\
        	\subfigure[(b) VideoGraph w/Video only (F-Score: 62.1)]{\includegraphics[width=1\linewidth]{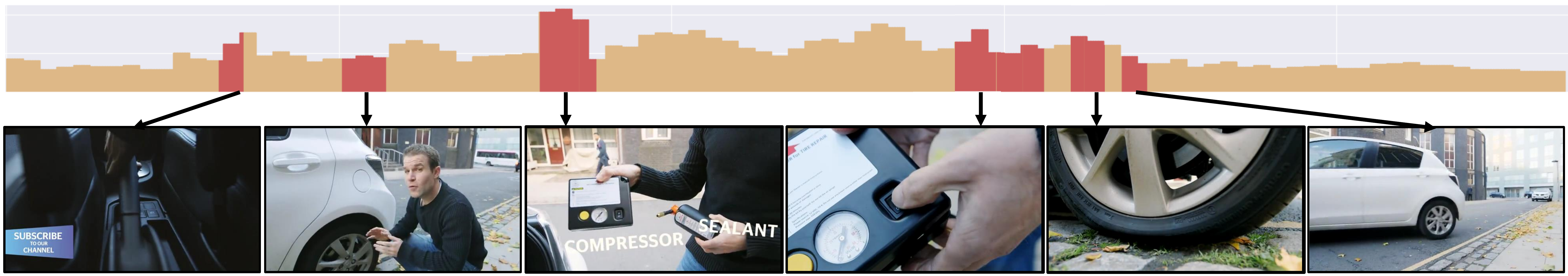}}\hfill\\
        	\subfigure[(c) VideoGraph w/Word-level query (F-Score: 70.2)]{\includegraphics[width=1\linewidth]{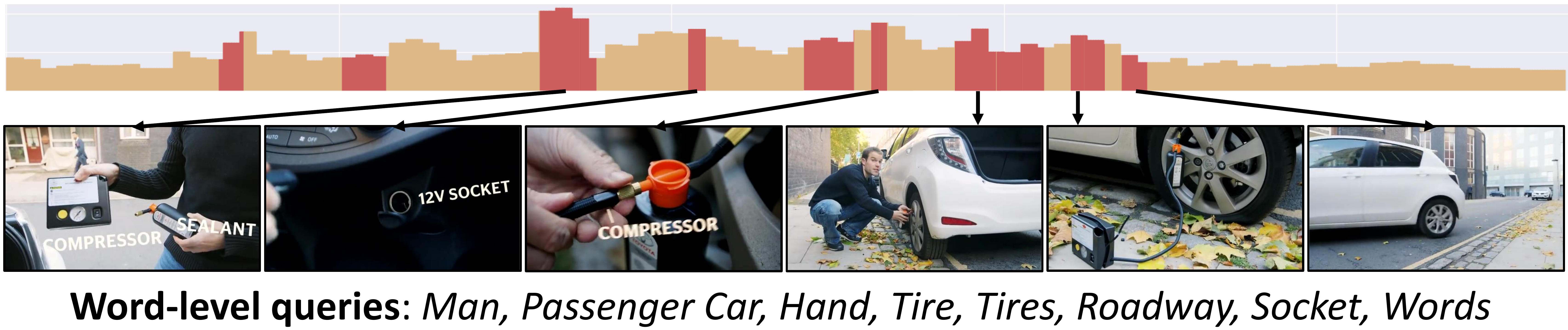}}\hfill
        	\caption{
        	Qualitative results on Video 2 from the TVSum benchmark~\cite{Song15}: (a) SumGraph~\cite{park20}, (b) VideoGraph without the language query, and (c) VideoGraph with the word-level language queries.
        	Brown bars show frame-level user annotations and red bars are selected subset shots for the summary video. Best viewed in color.
        }\label{fig:qual-comparison}
     \end{figure}  

    \begin{figure}[!t]
    	\centering
        	\renewcommand{\thesubfigure}{}
        	\subfigure[(a) Input Video from QFVS (P01)]{\includegraphics[width=1\linewidth]{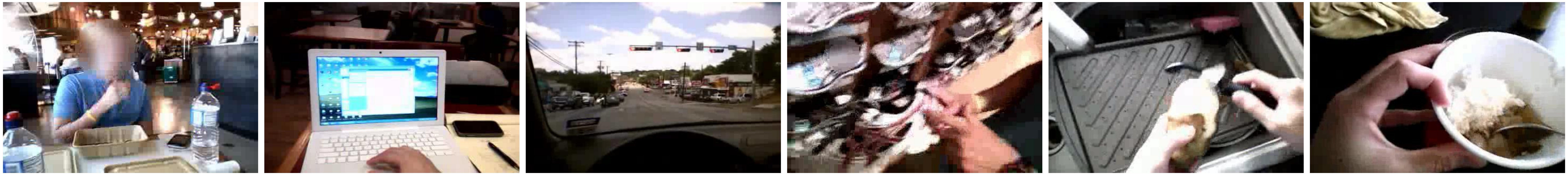}}\hfill\\
        	\subfigure[(b) Generic Summarization]{\includegraphics[width=1\linewidth]{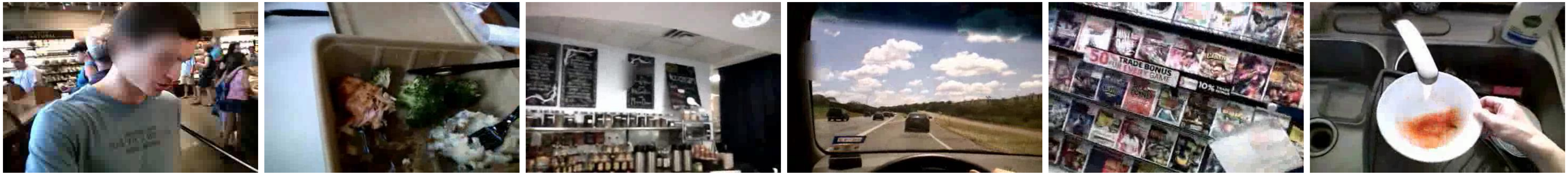}}\hfill\\
        	\subfigure[(c) Query-focused Summarization (Sky, Tree)]{\includegraphics[width=1\linewidth]{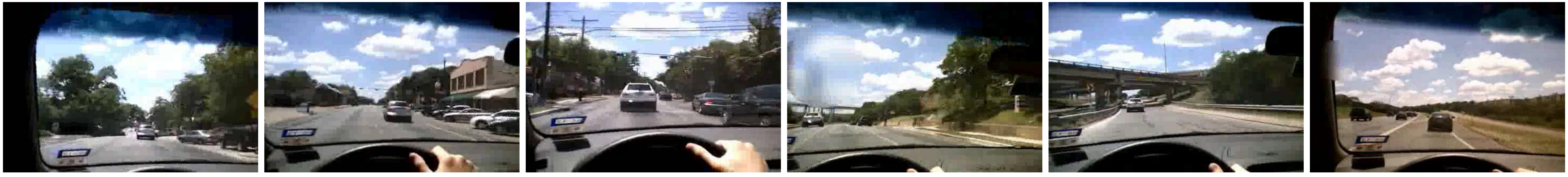}}\hfill   
         \\
       	\subfigure[(d) Query-focused Summarization (Food, Drink)]{\includegraphics[width=1\linewidth]{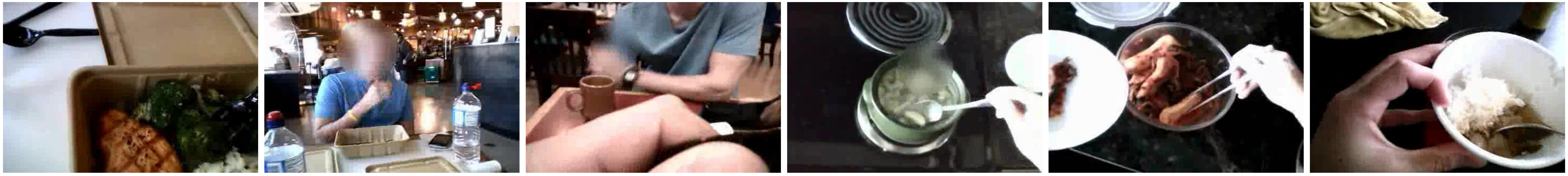}}\hfill\\
        	\caption{
        	Qualitative results on the QFVS benchmark~\cite{qfvs}: (a) input video, (b) generic summarization, (c) query-focused summarization with the queries ``Sky and Tree", and (d) query-focused summarization with the queries ``Food and Drink".
            We sample frames from the generated summaries.
        }\label{fig:qual-qfvs}
     \end{figure}
    
    \subsection{Qualitative Results}
    In \figref{fig:qual-comparison}, we provide qualitative results for the challenging video comprising visually similar frames.
    To visualize the summaries, we sampled frames from the keyshots selected by \cite{park20}, VideoGraph without the language queries, and with the word-level language queries.
    The main differences between the first two methods are the use of fine-grained frame features (i.e., object feature representations) and the SRR network.
    The summary generated by VideoGraph without the language queries captures almost all of the peak regions of the groundtruth scores, and the performance is also improved by $6.8\%$ in terms of F-score on the example video.
    This observation indicates that leveraging fine-grained frame features and considering object relationships enable more informative and meaningful summaries to be generated.
    The word-level language queries with VideoGraph derive an additional $8.1\%$ performance improvement, demonstrating the effectiveness of the language-guided video summarization framework.

     \begin{figure*}[!t]
        \centering
        {\includegraphics[width = 1.0\linewidth]{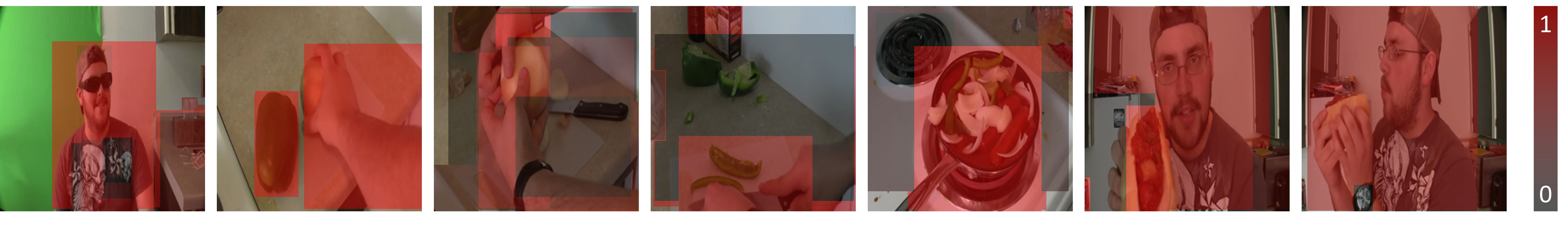}}
        \caption{
        Visualization for the dominance score of detected objects in the selected keyframes.
        }\label{fig:dominance}
     \end{figure*}

    \begin{figure}[!t]
    	\centering
        	\renewcommand{\thesubfigure}{}
        	\subfigure[(a) Input frame]{\includegraphics[width=0.5\linewidth]{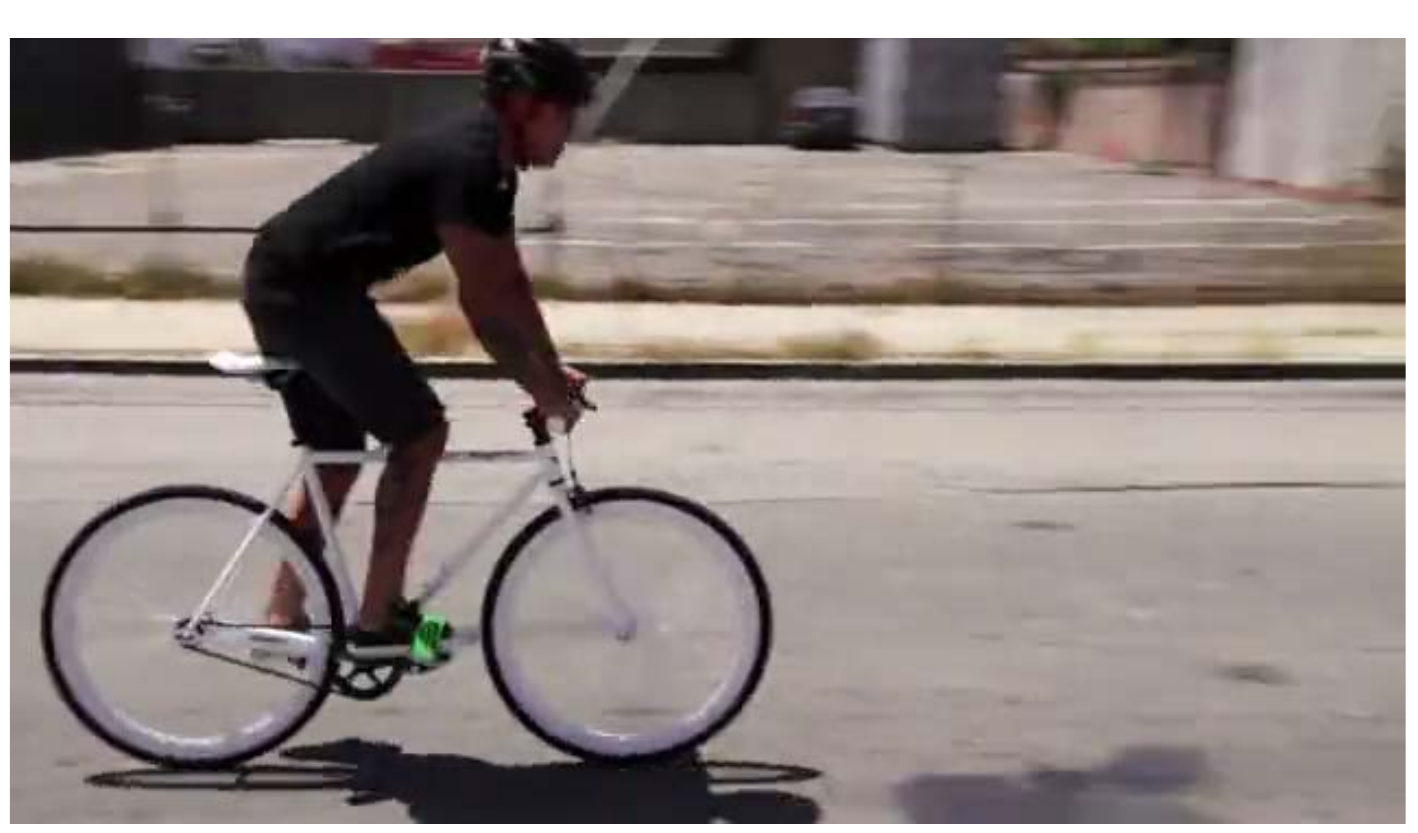}}\hfill
        	\subfigure[(b) $N=10$]{\includegraphics[width=0.5\linewidth]{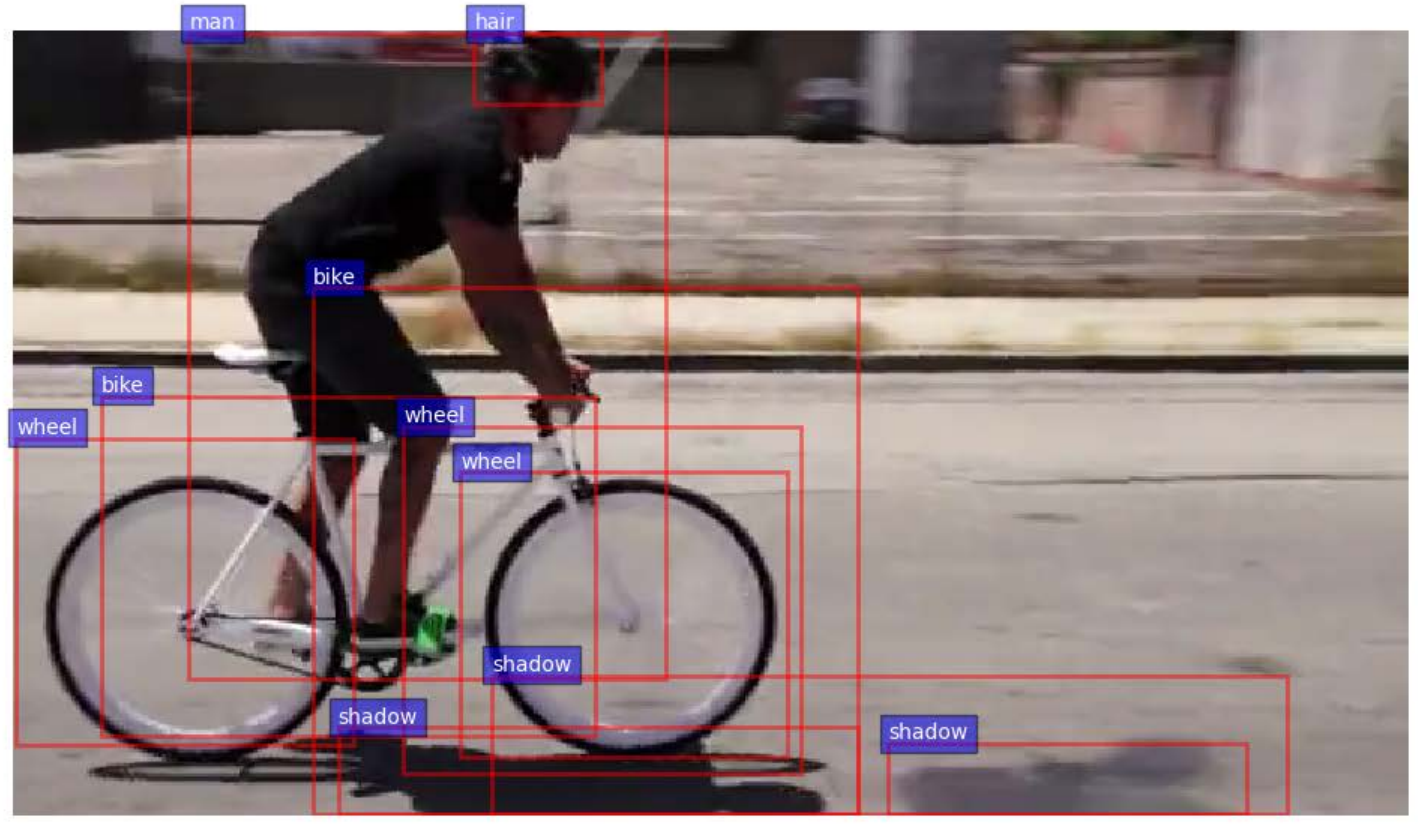}}\hfill\\
        	\subfigure[(c) $N=20$]{\includegraphics[width=0.5\linewidth]{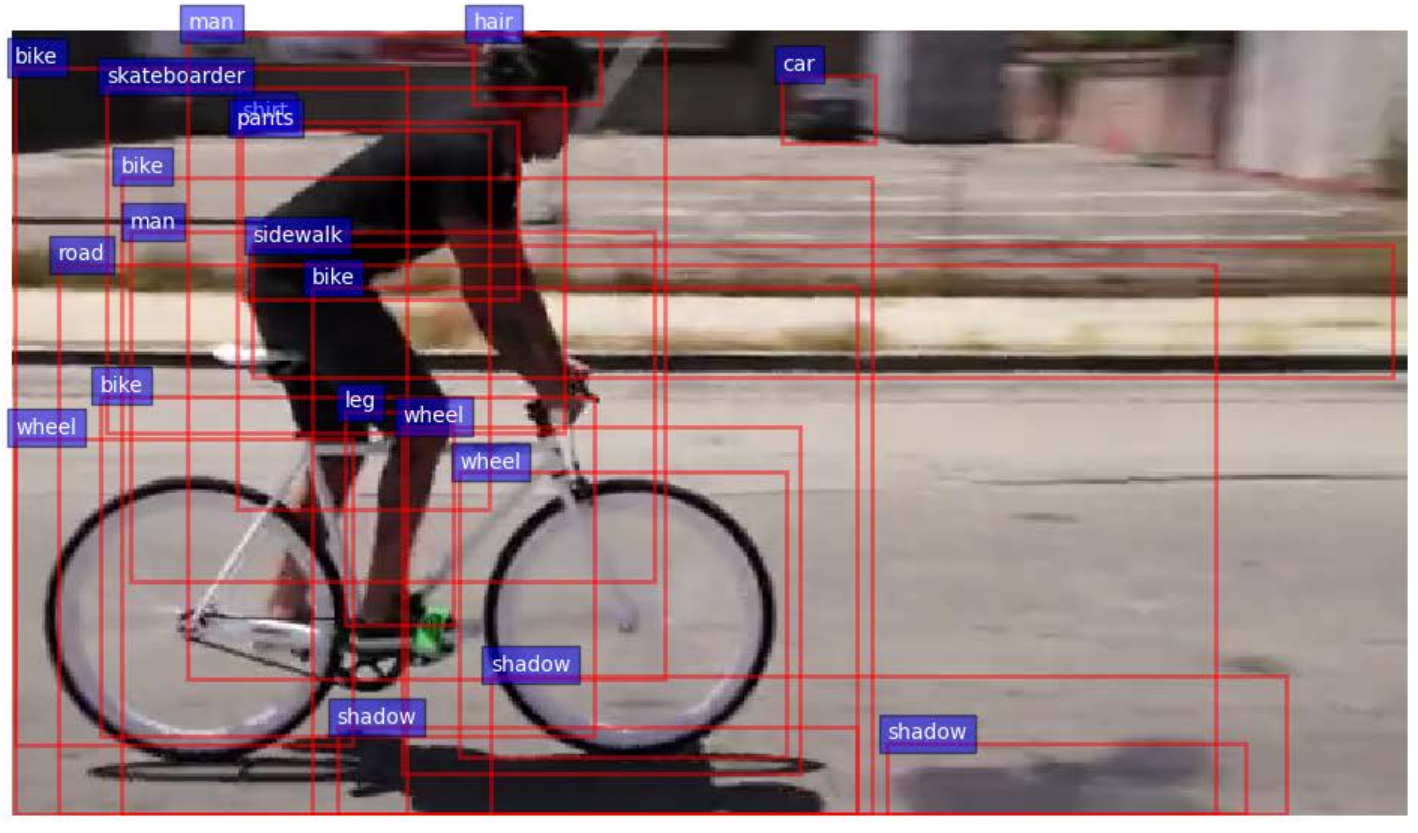}}\hfill
        	\subfigure[(d) $N=30$]{\includegraphics[width=0.5\linewidth]{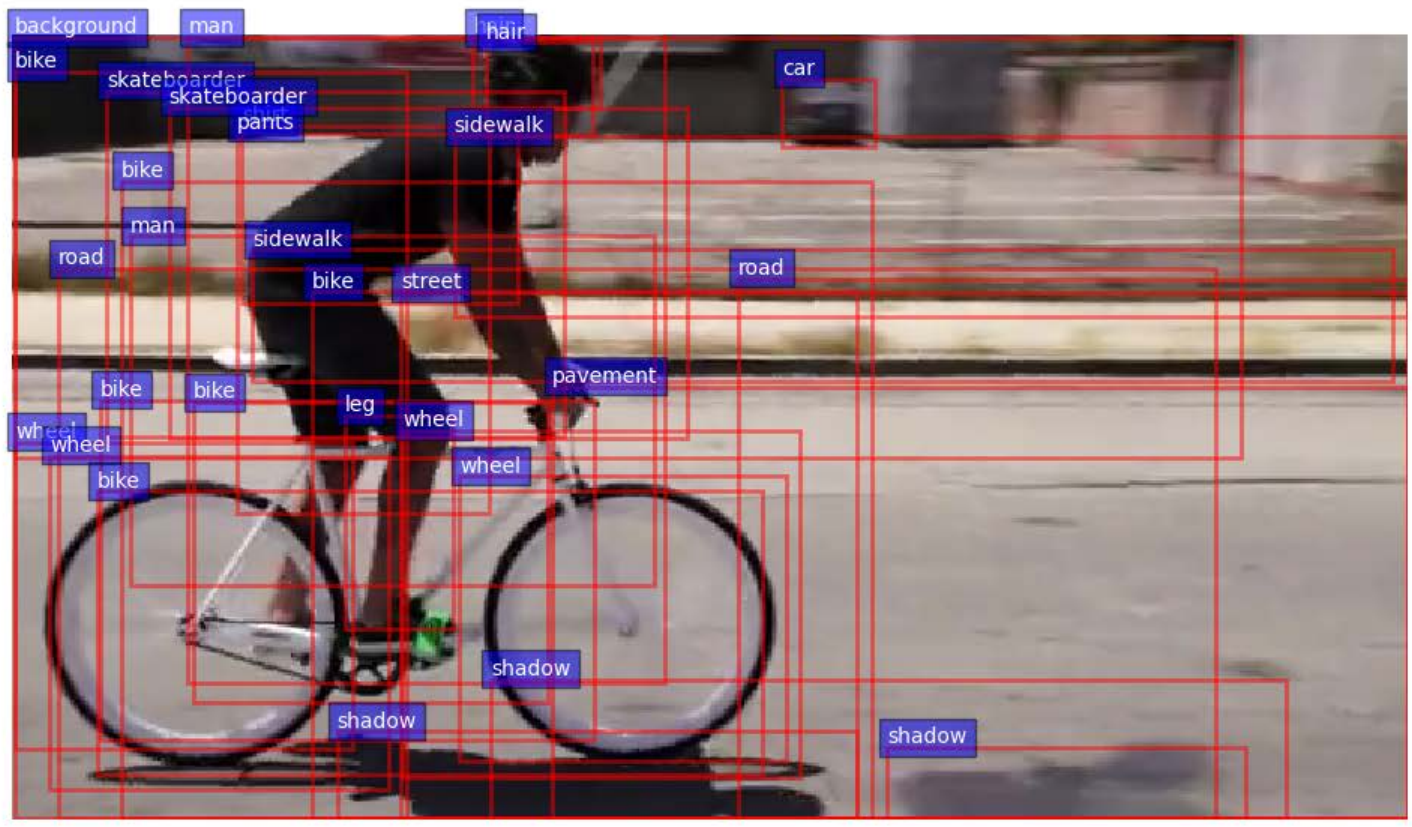}}\hfill\\
        	\caption{
        	(a) Input frame and (b)-(d) detected objects with respect to the number of objects $N$.
        	Noisy and redundant objects are detected as $N$ increases.
        }\label{fig:qual-objects}
    \end{figure}

    In \figref{fig:qual-qfvs}, we visualize the generic and query-focused summarization results on Video 1 from the QFVS dataset.
    The input video is about four hours long and contains multiple events related with several language queries.
    As shown in \figref{fig:qual-qfvs}-(b), the generic summarization, which is generated with the word-level language queries, contains various scenes that represent the whole video.
    \figref{fig:qual-qfvs}-(c) and (d) show that different summaries can be yield from the same video for the given different queries ``Sky and Tree" and ``Food and Drink".

    In video summarization, the importance of each object contained in a frame is not defined.
    We hypothesize that dominant objects contributing to the final summary video are highly correlated with large edge weights as object nodes propagate their information through the edges of the spatial graph in VideoGraph.
    Consequently, we define the dominance\footnote{We use the term ``dominance" instead of importance since ``importance score" is a predefined term that represents the importance of each frame in a video summarization task.} of each object as the normalized sum of its edge weights, representing the degree to which an object interacts with other objects within a frame.
    Specifically, we first aggregate the $n$-th object's edge weights:
    \begin{equation}
        \mathbf{d}_t = \lbrace d_{t, n} |  d_{t, n}= \sum_{i=1, i\neq n}^N {s_{t,ni}^K},  \quad n=1, ..., N \rbrace,
    \end{equation}
    where $s_{t, ni}^K$ is the edge weight between the $n$-th and $i$-th object nodes in the \textit{final} spatial graph of the $t$-th frame (i.e., $S_t^K$ in Equation~16 of the main paper).
    The dominance score $\bar{\mathbf{d}}_t$ can be computed by applying the min-max normalization to $\mathbf{d}_t$, such that,
    \begin{equation}
        \bar{\mathbf{d}}_{t} = \frac{\mathbf{d}_{t} - \min \mathbf{d}_{t}}{\max \mathbf{d}_{t} - \min \mathbf{d}_{t}}.
    \end{equation}
    We visualize the dominance score for objects within selected keyframes in \figref{fig:dominance}.
    The results show that objects pivotal to the video’s story tend to have higher dominance scores, as they are more strongly connected to other objects.
    This observation supports our hypothesis and suggests that VideoGraph effectively identifies and emphasizes information from key objects through the graph structure, enhancing summarization quality without manually assigning dominance scores.

    \begin{figure}[!t]
        \centering
        	\renewcommand{\thesubfigure}{}\hfill
        	\subfigure[(a) SumMe]{\includegraphics[width=0.48\linewidth]{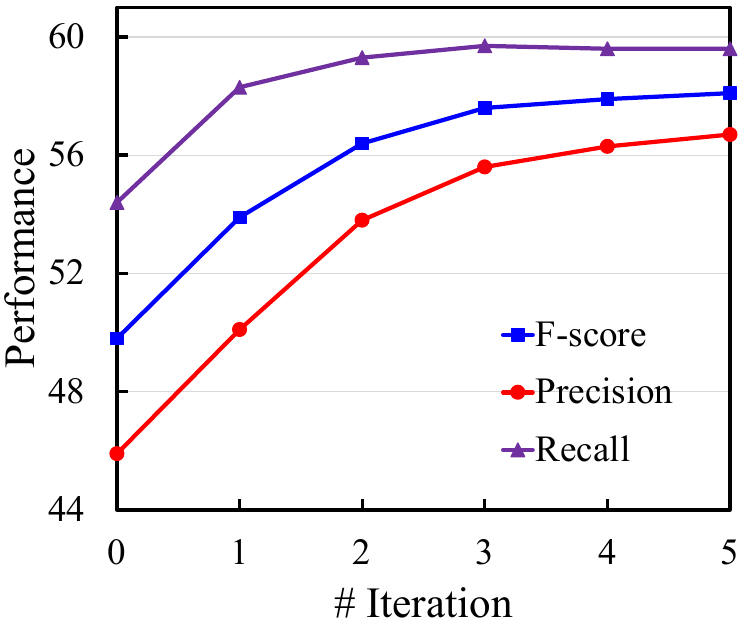}} \centering\hfill
        	\subfigure[(b) TVSum]{\includegraphics[width=0.48\linewidth]{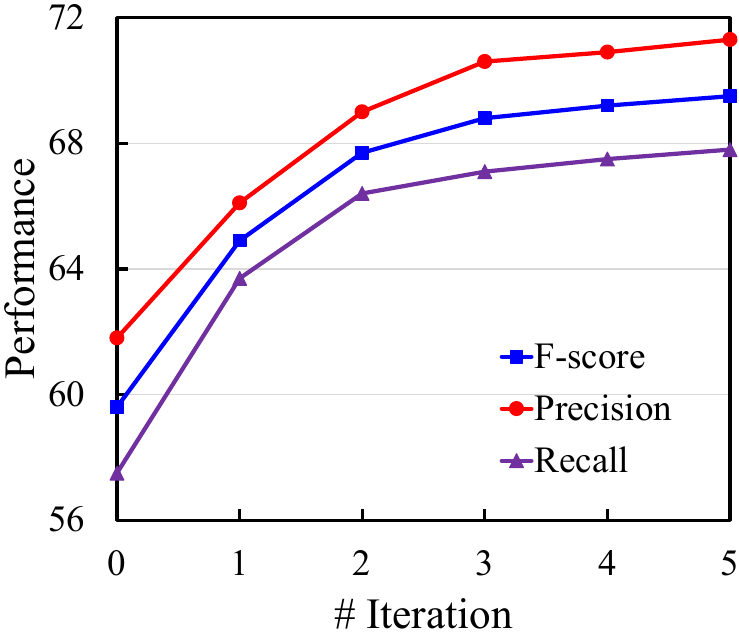}} \centering\hfill \vspace{-5pt}
        	\caption{
        	Convergence analysis of VideoGraph with respect to the number of iterations on (a) SumMe benchmark~\cite{Gygli14} and (b) TVSum~\cite{Song15} benchmarks.
        }\label{fig:8}
    \end{figure} 
    
    \subsection{Ablation Study}

    \begin{table*}[hbt!]
    \small
    \centering
    \caption{Ablation study for each component in VideoGraph on the SumMe~\cite{Gygli14}, TVSum~\cite{Song15}, and QFVS~\cite{qfvs} datasets.
    We report the F-score.
    }
    \begin{tabular}{cccccc}
        \toprule
        \multicolumn{2}{c}{Graph Topology}   &   \multirow{2}{*}{Language}  & \multirow{2}{*}{SumMe} &  \multirow{2}{*}{TVSum}  &   \multirow{2}{*}{QFVS} \tabularnewline
        \cmidrule{1-2}
        Temporal &   Spatial    &   &   &   &   \tabularnewline
        \midrule
        \cmark  &   \xmark  &   \xmark   & 51.4 & 63.9 &   - \tabularnewline
        \cmark  &   \xmark  &   \cmark & 55.8 & 67.1 &   67.27 \tabularnewline
        \xmark  &   \cmark  &   \xmark & 45.2 & 55.6 &   -        \tabularnewline
        \xmark  &   \cmark  &   \cmark   & 47.6 & 58.9 &   57.33 \tabularnewline
        \cmark  &   \cmark  &   \xmark   & 54.7 & 65.2 &  -    \tabularnewline
        \cmark  &   \cmark  &   \cmark & $\mathbf{58.1}$ & $\mathbf{69.5}$ &  $\mathbf{71.29}$ \tabularnewline
        \bottomrule
        \end{tabular}
    \label{tab:ablation_comp}
    \end{table*}
     
    \subsubsection{Component analysis}
    We analyze the contribution corresponding to the graph topology and language query of the proposed method on the SumMe, TVSum, and QFVS datasets, as shown in \tabref{tab:ablation_comp}.
    Note that the first row of the table corresponds to the SumGraph~\cite{park20}.
    The results show that the language query (i.e., sentence-level query) improves the performance of the SumGraph on all three datasets.
    On the other hand, the spatial graph topology without the temporal graph significantly degrades the performance, demonstrating the lack of temporal modeling capacity of the spatial graph topology.
    While the language query improves the performance even with the spatial graph, the performance is still insufficient compared to the model with the temporal graph.
    The full model of VideoGraph achieves significant performance improvement, consistently outperforming other baselines by large margins on three datasets.

    \begin{table}[!t]
    \centering
    \caption{
    Ablation study for the various combinations of loss functions in VideoGraph on the TVSum benchmark~\cite{Song15}.
    }
    \begin{tabular}{
    c c c c }
        \toprule
        Classification & Sparsity & \makecell{Diversity + \\Reconstruction} &  F-Score \tabularnewline
        \midrule
        \cmark & & & 65.4 \tabularnewline
        \cmark & \cmark &  & 67.8 \tabularnewline
        \cmark &  & \cmark & 67.5 \tabularnewline
        \cmark & \cmark & \cmark & \textbf{69.5} \tabularnewline
        \bottomrule
        \end{tabular}
        \label{tab:6}
    \end{table}

    \subsubsection{Loss functions}
    We also investigate the effectiveness of the loss functions.
    The four loss terms, (i.e., classification, sparsity, reconstruction, and diversity losses), are employed for supervised learning.
    While the reconstruction loss forces the reconstructed features to be similar to the original features, the diversity loss enforces the diversity of the reconstructed features.
    Therefore, we analyze the effectiveness of the reconstruction and diversity losses together.
    As shown in \tabref{tab:6}, the sparsity loss and the sum of the diversity and reconstruction losses improve the performance by $2.4\%$ and $2.1\%$, respectively, on the TVSum benchmark.
    The full use of loss functions contributes to a performance improvement of $4.1\%$.

    \begin{table}[!t]
    \centering
    \caption{
    Performance comparison evaluated on the TVSum benchmark~\cite{Song15} corresponding to the number of objects used in the SRR network.
    }
    \begin{tabular}{lccc}
        \toprule
        \# objects & Precision &  Recall &  F-score  \tabularnewline
        \midrule
        1   &   60.7   &   59.7   & 60.2   \tabularnewline
        2   &   61.7   &   60.8   & 61.2   \tabularnewline
        4   &   65.7   &   64.8   & 65.2 \tabularnewline
        8   &   68.5   &   65.3   & 66.9 \tabularnewline
        16  &   \textbf{71.3}   &   \textbf{67.8}   & \textbf{69.5}  \tabularnewline
        20  &   70.7   &   66.4   & 68.5  \tabularnewline
        36  &   64.1   &   64.4   & 66.2  \tabularnewline
        \bottomrule
        \end{tabular}
        \label{tab:num_object}
    \end{table}

    \begin{table*}[!t]
    \centering
    \caption{
    Ablation study for the number of the word-level language queries in VideoGraph on the SumMe~\cite{Gygli14}, TVSum~\cite{Song15}, and QFVS~\cite{qfvs} benchmarks.
    }
    \begin{tabular}{l c c c c c c c}
    \toprule
        \multirow{2}{*}[-0.3em]{Dataset} & \multicolumn{7}{c}{\# of words}  \tabularnewline
        \cmidrule{2-8}
            &   1   &   2   &   4   &   8   &   10  &   25  &   30  \tabularnewline
        \midrule
        SumMe &  54.6   &  55.8   & 57.2  & \textbf{58.1}  &    57.8  &  56.1    &   55.9    \tabularnewline
        TVSum &  65.4   &  66.3 &  67.9 & \textbf{69.5}  & 68.7    &    68.2    &   68.0    \tabularnewline
        QFVS    &  65.1 &   65.4    &   66.4    &   67.9    &   68.3    &    \textbf{68.8}    &   68.5\tabularnewline
        % QFVS &     &   &   &   &    \textbf{68.75} \tabularnewline
        \bottomrule
        \end{tabular}
        \label{tab:words}
    \end{table*}
    
    \subsubsection{Number of objects}
    We evaluated our model according to the number of objects used in the spatial graph.
    Following the original method~\cite{bottom-up}, we extract the top 36 confident region features in each frame and select $N$ objects in the order of the confidence score.
    The value of $N$ is set to $1$, $2$, $4$, $8$, $16$, $20$, and $36$.
    Note that, when $N$ is $1$, the SRR network is not used so that the network structure is the same as in our previous work~\cite{park20}.
    The only difference is that the input feature is the most salient region feature, not the GoogleNet~\cite{szegedy2015going} feature for the entire frame.
    As shown in \tabref{tab:num_object}, the performance increases as $N$ increases and decreases as $N$ becomes greater than $16$.
    Not surprisingly, the performance of $N=1$, $2$, and $4$ shows lower performance than our previous work~\cite{park20} since a small value of $N$ does not include the entire frame information.
    On the other hand, a large value of $N$ significantly degrades the summarization performance (e.g., $N=20$ and $36$).
    Since the feature extractor~\cite{bottom-up} is pretrained on the Visual Genome~\cite{visual-genomes} dataset containing a relatively large number of annotations per image, the features for many objects might contain redundant and noisy information.
    To validate this, we depict the detection results according to $N$ for the video frame from the TVSum~\cite{Song15} dataset in \figref{fig:qual-objects}.
    Based on this result, subsequent ablation studies are investigated by setting the number of objects $N$ to $16$ with the standard supervision configuration.

    \begin{figure}[!t]
    	\centering
        	\renewcommand{\thesubfigure}{}
        	\subfigure[(a) Iteration 0 (F-score: 45.2)]{\includegraphics[width=1\linewidth]{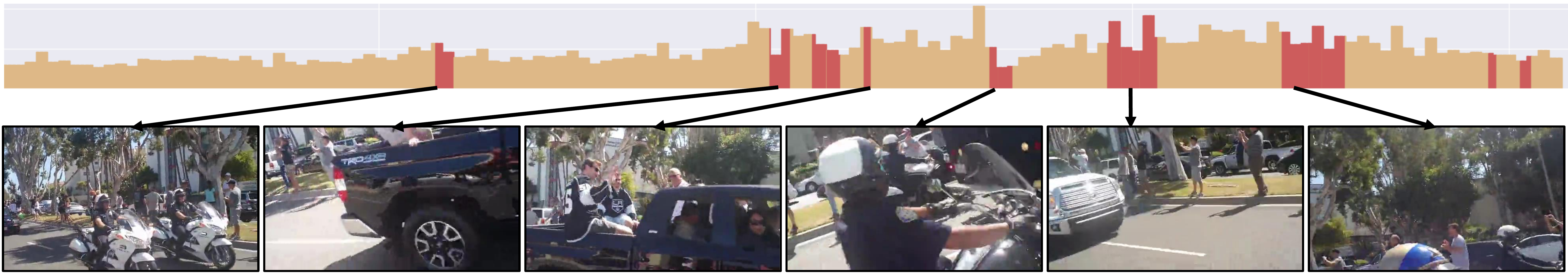}}\hfill\\
        	\subfigure[(b) Iteration 1 (F-score: 53.4)]{\includegraphics[width=1\linewidth]{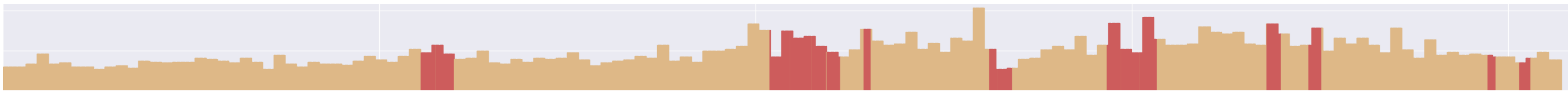}}\hfill\\
        	\subfigure[(c) Iteration 2 (F-score: 56.5)]{\includegraphics[width=1\linewidth]{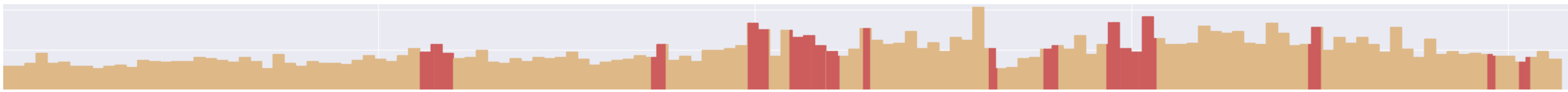}}\hfill\\
        	\subfigure[(d) Iteration 3 (F-score: 59.5)]{\includegraphics[width=1\linewidth]{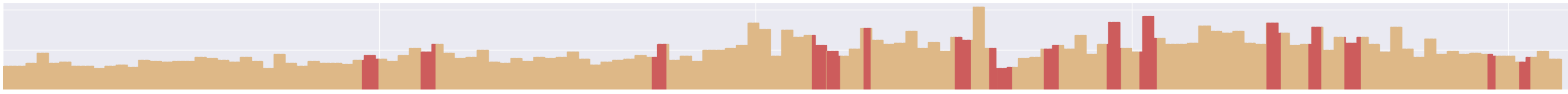}}\hfill\\
        	\subfigure[(e) Iteration 4 (F-score: 62.1)]{\includegraphics[width=1\linewidth]{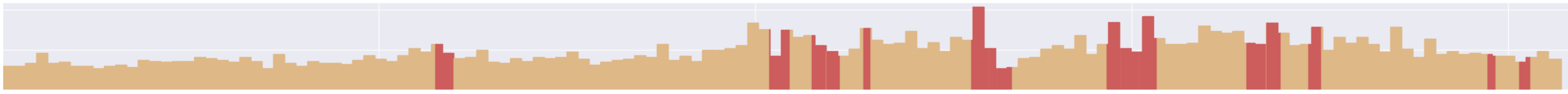}}\hfill\\
        	\subfigure[(f) Iteration 5 (F-score: 63.7)]{\includegraphics[width=1\linewidth]{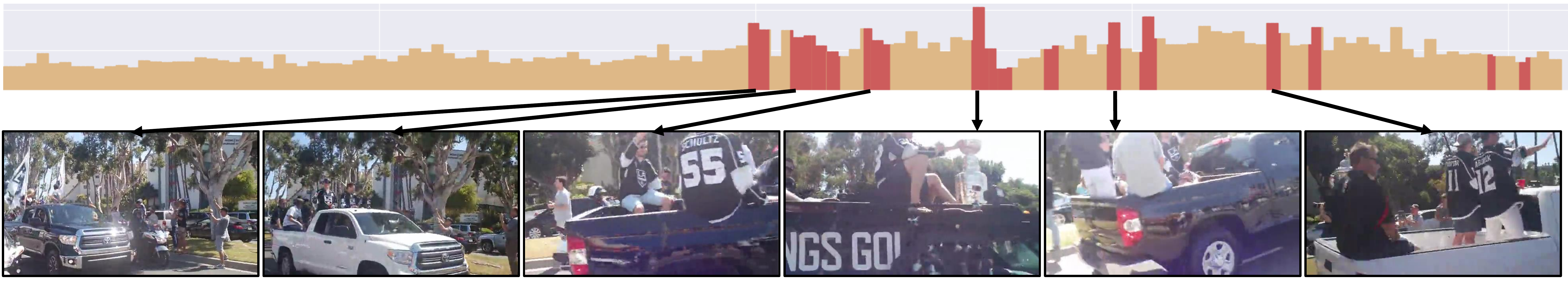}}\hfill \vspace{-5pt}
        	\caption{
        	Qualitative results for Video 28 on the TVSum benchmark~\cite{Song15} with respect to the number of iterations. Brown bars show frame-level user annotations, and red bars are selected subset shots for the summary.
        }\label{fig:qual-iteration}
     \end{figure}

    \subsubsection{Number of words}
    We analyze the effect according to the number of words, $W$, used as word-level language queries by measuring the generic video summarization performance on the SumMe, TVSum, and QFVS datasets.
    As shown in \tabref{tab:words}, the performance consistently increases until the number of words is 8 on the SumMe and TVSum datasets and decreases after that.
    The results validate that a small number of keywords can describe the videos in datasets and a larger number of language queries provide noisy information.

    Meanwhile, we attain the best performance on QFVS with a much larger number of language queries (i.e., $W=25$) than SumMe and TVSum, implying that a large number of keywords are required to describe long videos containing multiple events.
    In the main experiments, we thus use 8 words on SumMe and TVSum, and 25 words on QFVS, respectively.
    
    \begin{figure}[!t]
        	\centering
        	\renewcommand{\thesubfigure}{}
        	\subfigure[(a) Video 7]{\includegraphics[width=1.0\linewidth]{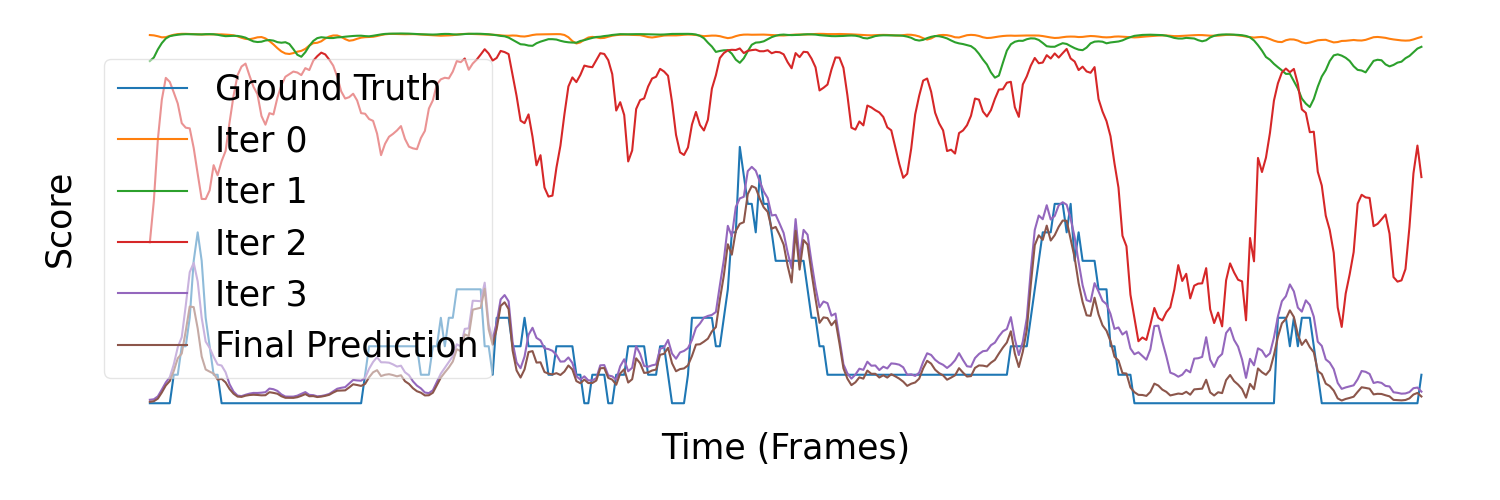}}\hfill \\
        	\subfigure[(b) Video 9]{\includegraphics[width=1.0\linewidth]{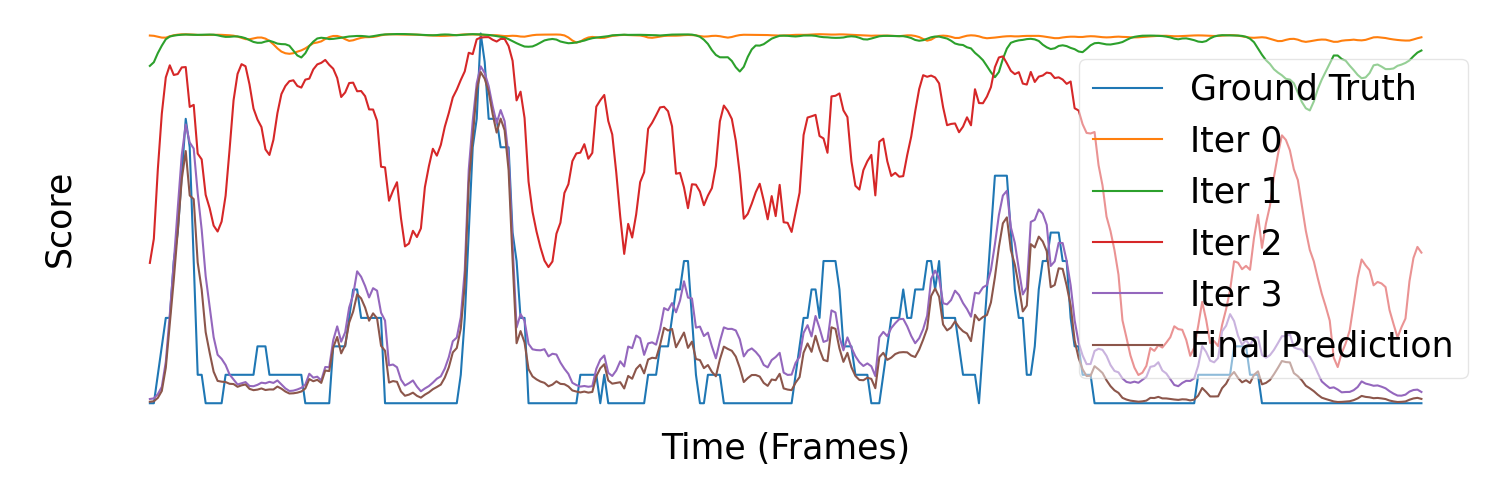}}\hfill \\
        	\subfigure[(c) Video 15]{\includegraphics[width=1.0\linewidth]{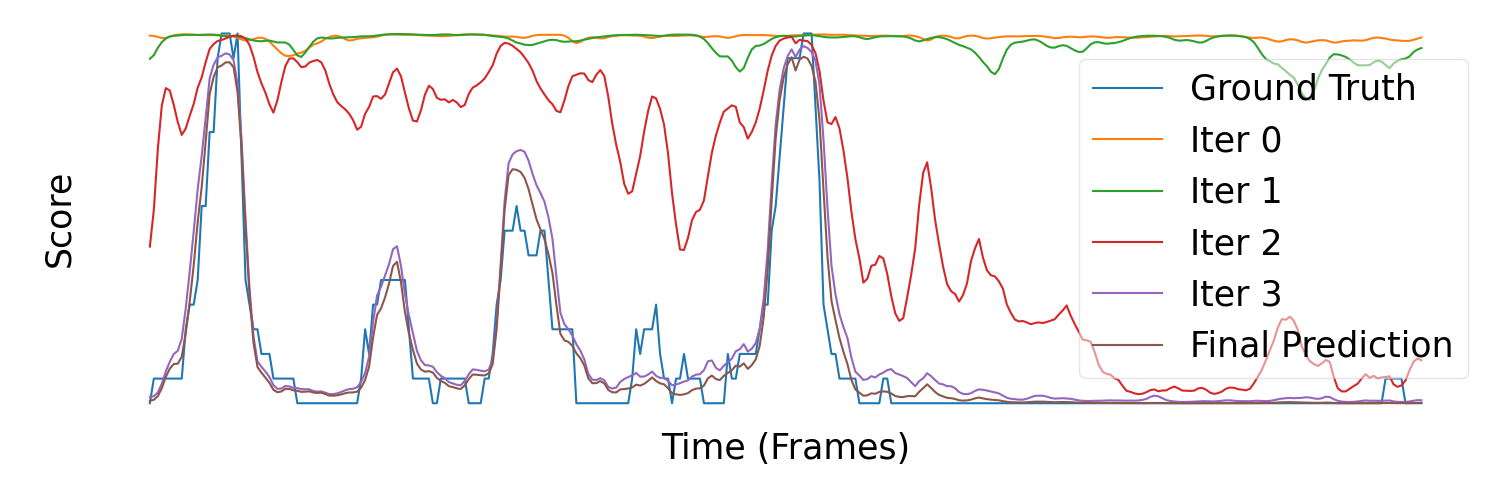}}\hfill 
        	\caption{
     	Predicted importance scores according to refinement iteration for videos from the TVSum benchmark~\cite{Song15}.
        }\label{fig:is_iteration}
    \end{figure} 

    \subsubsection{Number of iterations}
    To validate the effectiveness of the recursive graph refinement procedure used in VideoGraph, we examine the F-score corresponding to the number of iterations, as shown in \figref{fig:8}.
    Specifically, VideoGraph with the fixed initial graph (iteration 0) shows the performance of $49.8\%$ and $59.6\%$ in terms of F-score on the SumMe and TVSum datasets, respectively.
    The performance is gradually improved as the number of iterations increases, achieving $58.1\%$ and $69.5\%$ with 5 iterations on two datasets, and converging in three to five iterations.
    \figref{fig:qual-iteration} illustrates the groundtruth importance scores and keyframes selected by VideoGraph corresponding to the number of iterations.
    The selected keyframes are marked as red bars and the backgrounds are indicated as brown bars.
    The results show that VideoGraph selects frames with higher importance scores as the number of iterations increases.

    We further analyze the importance scores predicted by VideoGraph at each iteration, as shown in \figref{fig:is_iteration}.
    The results emphasize the importance of our graph refinement mechanism for effectively capturing semantic relationships between frames.
    Since the predictions in the initial graph (yellow lines) primarily rely on visual similarity, importance scores are relatively uniform across frames.
    However, as refinement iterations progress, the graph learns to establish meaningful semantic connections (i.e., discrimination for keyframes from non-keyframes).
    
    \subsubsection{Memory efficiency and computational cost}
    To investigate the efficiency of our model, we compare the total number of parameters, number of learnable parameters, runtime corresponding to each process for inference, and performance evaluated on SumMe~\cite{Gygli14} with the CNNs-based~\cite{Rochan18, park20} and transformer-based methods~\cite{clipit, mfst} in \tabref{tab:5}.
    {We measure runtimes using a single 24GB TITAN RTX GPU.
    % The runtime for only summarization is measured with pre-extracted features and the total runtime includes caption generation, feature extraction, and summarization.
    We repeat the inference 300 times for all videos, which have 320 frames, and record the average value as runtime.
    We follow the procedure of the official implementation of BMT~\cite{captionmodel}\footnote{Official code: \url{https://github.com/v-iashin/BMT}} to generate video captions.
    For GoogleNet~\cite{szegedy2015going} and Faster R-CNN~\cite{faster-rcnn}, we use open-source models provided by torchvision library and Detectron2 repository\footnote{Official code: \url{https://github.com/facebookresearch/detectron2}}, respectively.
    For CLIP image and text encoders~\cite{clip}, we use the CLIP ViT-B/32 pretrained on LAION-400M~\cite{laion400m}\footnote{Official code: \url{https://huggingface.co/docs/transformers/model_doc/clip}}.}
    % We measure each component according to the use of the language query (i.e., w/o Language query, w/Sentence-level query, and w/Word-level query).
    For the baseline network, we used $16$ object features, three graph convolution layers each for the SRR and TRR networks, and two graph convolution layers for the summarization network.
    The number of iterations for graph refinement is set to five.

    \begin{table*}[!t]
    \centering
    \small
        \caption{
        Use of language queries (Lan.) and captioning model (Cap. model), total number of parameters (Total params.), number of learnable parameters (Learn. params.), and runtimes for caption generation (Cap.), feature extraction (FE), and summarization (Sum.).
        }
        \begin{tabular}{lcccccccc}
        \toprule
            \multirow{2}{*}{Method} & \multirow{2}{*}{Lang.}   &    \multirow{2}{*}{\makecell{Cap.\\model}} &   \multirow{2}{*}{\makecell{Total \\ params.}}& \multirow{2}{*}{\makecell{Learn.\\ params.}} &  \multicolumn{3}{c}{Runtime} &    \multirow{2}{*}{\makecell{SumMe \\(F-score)}}\tabularnewline   \cmidrule{6-8}
            &   &   &   &   &  {Cap.} &  {FE}   &  {Sum.}    &   \\
            \midrule
            Rochan et al.~\cite{Rochan18} & \xmark  &  \xmark  & 123.1M   & 116.5M &    -   &  129ms & 84ms &  47.5    \tabularnewline
            Park et al.~\cite{park20}  &    \xmark & \xmark  &    12.1M & 5.5M &    -   &   129ms   & 13ms &   51.4    \tabularnewline
            Narasimhan et al.~\cite{clipit} & \cmark    & \cmark &    205.6M   &   5.8M &    250s  &   317ms   &   23ms  &   54.2    \tabularnewline
            Park et al.~\cite{mfst} w/o Audio  & \cmark  & \cmark  &   280.1M   &   31.7M    &   250s  &   1057ms &   69ms  &   54.2    \tabularnewline
            \midrule            
            \textbf{VideoGraph} w/Video only  &   \xmark  &   \xmark  &    49.3M    & {4.7M} & -  &   474ms   &   {11ms} &    54.7    \tabularnewline
            \textbf{VideoGraph} w/Sentence    &   \cmark  &   \cmark  &   176.3M   & {17.7M} &    250s  &   485ms &    {32ms} &   57.0   \tabularnewline
            \textbf{VideoGraph} w/Word    &   \cmark  &   \xmark    &  124.3M   & {16.7M} & - &   484ms &   {31ms} &   \textbf{58.1}    \tabularnewline
            \bottomrule
            \end{tabular}
        \label{tab:5}
    \end{table*}
    
    \noindent\textbf{For only summarization with pre-extracted features.} For inference, while \cite{Rochan18} requires approximately 84ms using about 116.5M network parameters for a video containing 320 frames, our model without the language query takes 11ms using about 4.7M network parameters under the same settings.
    This result indicates that VideoGraph surpasses \cite{Rochan18} in terms of memory efficiency and summarization performance.
    Our model uses 0.8M fewer parameters and takes 2ms less time than \cite{park20}, with $2.3\%$ and $1.3\%$ performance improvement on the SumMe and TVSum datasets, respectively, demonstrating the effectiveness of learning fine-grained relationships between frames.
    
    On the other hand, a large increase in the size of backbone networks is remarkable in comparison with the transformer-based methods~\cite{clipit, mfst}.
    Specifically, they employed the CLIP ViT-B/32 image and text encoders~\cite{clip}, which have 85.8M and 63.0M parameters, for visual and language representations, respectively.
    For caption generation, they identically used the bi-modal transformer (BMT)~\cite{captionmodel}.
    Our VideoGraph achieves better performance than \cite{clipit}, showing compatible runtime and the number of learnable parameters.
    
    \cite{mfst} additionally exploited the VideoCLIP~\cite{videoclip} for visual representations, and they learn a large number of trainable parameters, showing a low memory efficiency and taking about 2.2x more runtime than our VideoGraph for inference.
    
    The comparison between our three baselines shows that most of the computation burden is concentrated on the transformers.
    Although the model with the language query takes about $3\times$ more parameters and runtime than the model without the language query, it is worth noting that considerable performance improvements and customized summaries can be obtained with the language-guided model.

    \noindent\textbf{For the whole inference from scratch.}
    As shown in \tabref{tab:5}, we notice that most of the time for inference is concentrated on caption generation, taking about 4 minutes on average.
    In this regard, it is worth noting that VideoGraph with word-level queries incorporates language information into video summarization without caption generation and achieves even better performance.
    Although visual feature extraction is the second longest part, it takes relatively less time than caption generation.
    In addition, GoogleNet, which is the lightest visual encoder, takes 129ms for a video, and the transformer-based backbones, i.e., CLIP~\cite{clip} and VideoCLIP~\cite{videoclip}, require 306ms and 740ms, respectively.
    The Faster R-CNN~\cite{faster-rcnn} used in our VideoGraph takes 474ms for the video, which is comparable to other encoders.
    Consequently, VideoGraph with word-level language queries significantly outperforms the previous works while mitigating the burden of caption generation.

\section{Conclusion and Future Work}
    In this paper, we proposed VideoGraph, which formulates video summarization as a language-guided graph modeling problem.
    Contrary to previous works that employed global visual representations and sentence-level language queries, we leverage fine-grained object representations and their corresponding word-level language queries, ensuring consistent information between visual and language data.
    % In contrast to the previous works that applied convolution or recurrent operations, the proposed approach learned object-object and frame-frame semantic interactions by incorporating the transformers and graph convolutional networks.
    VideoGraph constructs a language-guided spatiotemporal graph through spatial and temporal relation networks, and recursively refines the graph to enable graph nodes to be gradually connected based on semantic relationships.
    % Furthermore, VideoGraph recursively estimates and refines initially constructed graphs to learn semantic relationships.
    % To leverage spatiotemporal graph convolutional networks (ST-GCNs) in a recursive manner, we separate ST-GCNs into spatial and temporal relation reasoning networks.
    We demonstrated the effectiveness of each component in VideoGraph, including graph topology, types of language queries, loss functions, and graph refinement, and achieved state-of-the-art performance in both the supervised and unsupervised settings.
    % We validated the effectiveness corresponding to the different types of language queries through extensive experiments.
    % In addition, the network was trained in both a supervised and an unsupervised manner, significantly improving the performance over state-of-the-art approaches.

    There are many possible directions for future work.
    While we achieved a new state-of-the-art on several datasets, we use off-the-shelf pretrained models~\cite{captionmodel,faster-rcnn,clip} to generate and encode the language queries.
    This could allow inappropriate biases from the pretrained models to be propagated to our model.
    Moreover, the pretraining dataset of the object detector for the word-level language queries should also be carefully regarded to generate more specific summaries.
    In this regard, our further works will be focused on generalizing the proposed approach to deal with robust summarization performance on language queries generated from pretrained models.

\section*{Data Availability Statement}
    The SumMe and QFVS datasets were publicly available on \url{https://gyglim.github.io/me/vsum/index.html} and \url{https://www.aidean-sharghi.com/cvpr2017}, respectively.
    However, we found that they are inaccessible now so we will provide pre-extracted features upon reasonable request.
    The TVSum dataset is publicly available on \url{https://github.com/yalesong/tvsum}.

\section*{Acknowledgement}
    This work was supported by the National Research Foundation of Korea (NRF) grant funded by the Korea government (MSIT) (No.~RS-2025-00515741 and No.~RS-2025-02216328) and the Yonsei Signature Research Cluster Program of 2024 (2024-22-0161).
    \bibliography{egbib}% common bib file

\end{document}